\documentclass{article}
\usepackage{ijcai25}

\usepackage{times}
\usepackage{soul}
\usepackage{url}

\usepackage[colorlinks=true, linkcolor=blue, citecolor=blue, urlcolor=blue]{hyperref}

\usepackage[utf8]{inputenc}
\usepackage[small]{caption}
\usepackage{graphicx}
\usepackage{amsmath}
\usepackage{amsthm}
\usepackage{booktabs}
\usepackage{tcolorbox}
\usepackage[ruled,linesnumbered]{algorithm2e}
\usepackage[switch]{lineno}
\usepackage{enumerate}
\usepackage{multirow}
\usepackage{subfig}
\usepackage{color}
\usepackage{amssymb} 
\usepackage{placeins}
\usepackage{listings}
\usepackage{xcolor}
\usepackage{tcolorbox}
\usepackage{fontawesome}
\usepackage{amssymb}
\tcbuselibrary{listings, skins}
\definecolor{codegreen}{rgb}{0,0.6,0}
\definecolor{codegray}{rgb}{0.5,0.5,0.5}
\definecolor{codepurple}{rgb}{0.58,0,0.82}
\definecolor{backcolour}{rgb}{0.95,0.95,0.92}

\newtcblisting{mypython}[1][]{
    listing only,
    listing options={
        style=tcblatex,
        commentstyle=\color{codegreen},       % 注释样式
        keywordstyle=\color{magenta},         % 关键字样式
        numberstyle=\tiny\color{codegray},    % 行号样式
        stringstyle=\color{codepurple},       % 字符串样式
        basicstyle=\ttfamily\footnotesize,           % 基本字体样式
        breakatwhitespace=false,              % 在空格处断行
        breaklines=true,                      % 自动换行
        captionpos=b,                         % 标题位置（底部）
        keepspaces=true,                      % 保留空格
        numbersep=5pt,                        % 行号与代码的间距
        showspaces=false,                     % 不显示空格
        showstringspaces=false,               % 不显示字符串中的空格
        showtabs=false,                       % 不显示制表符
        tabsize=4,                            % 制表符宽度
        language=Python,       
    },
    colback=gray!10, % 背景颜色
    colframe=black,     % 边框颜色
    arc=5pt,            % 圆角半径
    boxrule=1.5pt,      % 边框宽度
    left=2mm,           % 左边距
    right=2mm,          % 右边距
    top=1mm,            % 上边距
    bottom=1mm,         % 下边距
}

\newtcblisting{mypython1}[1][]{
    listing only,
    listing options={
        style=tcblatex,
        commentstyle=\color{codegreen},       % 注释样式
        keywordstyle=\color{magenta},         % 关键字样式
        numberstyle=\tiny\color{codegray},    % 行号样式
        stringstyle=\color{codepurple},       % 字符串样式
        basicstyle=\ttfamily\tiny,           % 基本字体样式
        breakatwhitespace=false,              % 在空格处断行
        breaklines=true,                      % 自动换行
        captionpos=b,                         % 标题位置（底部）
        keepspaces=true,                      % 保留空格
        numbersep=5pt,                        % 行号与代码的间距
        showspaces=false,                     % 不显示空格
        showstringspaces=false,               % 不显示字符串中的空格
        showtabs=false,                       % 不显示制表符
        tabsize=2,                            % 制表符宽度
        language=Python,       
    },
    colback=gray!10, % 背景颜色
    colframe=black,     % 边框颜色
    arc=5pt,            % 圆角半径
    boxrule=1.5pt,      % 边框宽度
    left=0.5mm,           % 左边距
    right=0.5mm,          % 右边距
    top=0.2mm,            % 上边距
    bottom=0.2mm,         % 下边距
}

\newcommand{\blue}[1]{{\color{blue}{#1}}}

\title{Pseudocode-Injection Magic: Enabling LLMs to Tackle\\ Graph Computational Tasks}

\author{
    Author Name
    \affiliations
    Affiliation
    \emails
    email@example.com
}

\author{
Chang Gong\thanks{Equal contribution.}
\and
Wanrui Bian\footnotemark[1]\and
Zhijie Zhang\footnotemark[1]\And
Weiguo Zheng\thanks{Corresponding author.}\\
\affiliations
School of Data Science, Fudan University, Shanghai, China\\
\emails
\{changgong22,wrbian23,zhangzj22\}@m.fudan.edu.cn,
zhengweiguo@fudan.edu.cn
}

\begin{document}

\maketitle

\begin{abstract}
% Graph computational tasks are complex and challenging and usually require to design sophisticated algorithms to solve. With the rise of LLMs, researchers attempt to explore using LLMs to address these tasks. However, existing methods are constrained by LLM's insufficient ability of graph understanding and prohibitive inference costs, thus they cannot be used to process large-scale graphs. Inspired by how humans approach graph problems, we propose a new framework PIE (\underline{{P}}seudocode-\underline{{I}}njection-\underline{{E}}nhanced LLM Reasoning on Graph Computational Tasks), which consists of three steps: problem understanding, prompt design, and code generation and execution. Specifically, LLMs are responsible for understanding tasks and extracting relevant information to write correct code, while the task of analyzing the specific graph structure and executing code is assigned to the interpreter. To assist LLMs in generating efficient code, we further search task-related pseudocodes and inject them into prompt. We also employ cost-effective trials-and-errors to ensure LLM's output codes run correctly. Compared to other methods that call LLMs for each test case, PIE only calls LLMs during code generation and the output code can be reused, which reduces LLM inference cost. Extensive experiments show that PIE significantly outperforms other baselines in accuracy with the lowest computational cost. The code of our framework is publicly available at \href{https://github.com/hmtbgc/PIE}{https://github.com/hmtbgc/PIE}.

Graph computational tasks are inherently challenging and often demand the development of advanced algorithms for effective solutions. With the emergence of large language models (LLMs), researchers have begun investigating their potential to address these tasks. However, existing approaches are constrained by LLMs' limited capability to comprehend complex graph structures and their high inference costs, rendering them impractical for handling large-scale graphs.
Inspired by human approaches to graph problems, we introduce a novel framework, PIE (\underline{{P}}seudocode-\underline{{I}}njection-\underline{{E}}nhanced  LLM Reasoning for Graph Computational Tasks), which consists of three key steps: problem understanding, prompt design, and code generation. In this framework, LLMs are tasked with understanding the problem and extracting relevant information to generate correct code. The responsibility for analyzing the graph structure and executing the code is delegated to the interpreter. We inject task-related pseudocodes into the prompts to further assist the LLMs in generating efficient code. We also employ cost-effective trial-and-error techniques to ensure that the LLM-generated code executes correctly.
Unlike other methods that require invoking LLMs for each individual test case, PIE only calls the LLM during the code generation phase, allowing the generated code to be reused and significantly reducing inference costs. Extensive experiments demonstrate that PIE outperforms existing baselines in terms of both accuracy and computational efficiency. 

% The code for our framework is publicly available at  \href{https://github.com/hmtbgc/PIE}{https://github.com/hmtbgc/PIE}.

\end{abstract}

\section{Introduction}

Graph reasoning requires designing specialized algorithms tailored to the unique characteristics of each task. For instance, tasks like community detection and subgraph matching necessitate distinct approaches, making the development and implementation of such algorithms both resource-intensive and computationally expensive. Consequently, developing efficient and cost-effective solutions for diverse graph reasoning tasks remains a critical challenge in practical applications.

% In recent years, large language models (LLMs), such as GPT and its successors, have achieved remarkable advancements in natural language processing (NLP) and beyond~\cite{?}. These models, equipped with billions of parameters, have demonstrated impressive generalization capabilities, enabling them to address a wide range of problems across diverse domains. Leveraging extensive pretraining on large-scale and diverse datasets, LLMs can understand and generate human-like text, reason about complex queries, and perform tasks that extend far beyond traditional NLP, including code generation, mathematical reasoning, and scientific problem-solving.

% In recent years, large language models (LLMs)~\cite{gpt4,llama3,deepseek} have achieved remarkable advances in diverse domains. These models, equipped with billions of parameters, have demonstrated impressive generalization capabilities. After pretraining on large-scale and diverse datasets, LLMs can understand and generate human-like text, reason about complex queries, and perform tasks that extend far beyond traditional NLP, including code generation~\cite{deepseek-coder,qwen-coder,debugllm}, mathematical reasoning~\cite{deepseek-math,jiuzhang,InternLMMath}, and scientific problem-solving~\cite{pangu,funsearch,scibench}.

In recent years, large language models (LLMs)~\cite{gpt4,llama3,deepseek} have achieved remarkable advances across various domains. With billions of parameters, these models exhibit impressive generalization abilities. After pretraining on large-scale and diverse datasets, LLMs can understand and generate human-like text, reason through complex queries, and tackling tasks that extend beyond traditional natural language processing (NLP), including code generation~\cite{deepseek-coder,qwen-coder,debugllm}, mathematical reasoning~\cite{deepseek-math,jiuzhang,InternLMMath}, and scientific problem-solving~\cite{pangu,funsearch,scibench}.
%
% Building on their versatility, researchers have recently begun investigating the application of LLMs to graph-related tasks. A common approach involves directly assigning graph reasoning tasks to LLMs and relying on the model to produce solutions. By representing graph problems as natural language prompts or structured queries, these methods aim to enable LLMs to interpret the underlying task, reason about the graph, and return results without the need for explicit algorithmic implementations by humans. This emerging direction leverages the general-purpose reasoning capabilities of LLMs, opening new possibilities for addressing graph reasoning tasks in novel and flexible ways.
%
% Building on their versatility, researchers begin to investigate LLMs' application on graph computational tasks~\cite{grapharena,graphinstruct,graphteam,nlgraph,pse}. A common approach is assigning graph computational tasks to LLMs directly and asking them to give solutions. By representing graph problems as natural language prompts or structured queries, LLMs attempt to understand the task, perform reasoning on the graph, and return results without humans' help. This emerging direction leverages the general-purpose reasoning capabilities of LLMs, opening new possibilities for addressing graph computational tasks in novel and flexible ways.
%
%
Leveraging their versatility, researchers have begun exploring the application of large language models (LLMs) to graph computational tasks~\cite{grapharena,graphinstruct,graphteam,nlgraph,pse}. A common approach involves directly assigning graph problems to LLMs, presenting the tasks as natural language prompts or structured queries. The LLMs then attempt to understand the problem, perform reasoning on the graph, and generate solutions autonomously, without human intervention. This emerging approach capitalizes on the general-purpose reasoning abilities of LLMs, offering innovative and flexible solutions to graph computational challenges.
%
% However, while this strategy has shown potential, there are significant limitations. These limitations can be summarized as follows:
However, this strategy %still has 
suffers from 
two problems:

\begin{enumerate}[(1)]
    % \item \textbf{Accuracy limitations:} The accuracy of LLMs in solving graph reasoning tasks is often insufficient, particularly when dealing with complex or large-scale graphs. One major limitation stems from the context length constraints of LLMs, which can hinder their ability to fully encode the structure and relationships within large graphs. Furthermore, LLMs rely on probabilistic autoregressive decoding, inherently prioritizing plausible predictions over exact computations. This makes them prone to errors even in relatively simple tasks, such as counting the number of nodes, edges, or calculating the degree of a node.
    % For more demanding tasks, such as finding optimal paths, detecting communities, or solving combinatorial problems, these limitations often lead to suboptimal or incorrect solutions. Such unreliability becomes especially problematic in scenarios where correctness is critical.

    \item 
    %\textbf{Accuracy limitations:} The ability of LLMs to understand graph structure is insufficient, particularly for complex or large-scale graphs. Since graph structure is unordered, transforming it into a text sequence makes LLMs to fail to capture topological 
    %positional
   % and semantic relationships between nodes. Moreover, LLMs rely on probabilistic autoregressive decoding, inherently prioritizing plausible predictions over exact computations, which leads to prediction errors even in simple tasks. For example, we use llama3-8b to perform graph reading tasks including node and edge number, as well as maximum, minimum, and average degrees of nodes. Table~\ref{readgraph} shows that, except node number and maximum degree on small graphs, other graph information cannot be correctly identified. On large-scale graphs, LLM's performance is extremely poor. Thus, for more demanding tasks like finding optimal paths or solving combinatorial problems, it is more challenging for LLMs to provide correct solutions.
%
    \textbf{Accuracy Limitations:} The ability of LLMs to understand graph structures is limited, especially when dealing with complex or large-scale graphs. Since graph structures are unordered, transforming them into a text sequence prevents LLMs from effectively capturing the topological and semantic relationships between nodes. Moreover, LLMs rely on probabilistic autoregressive decoding, which inherently prioritizes plausible predictions over precise computations. This leads to prediction errors, even in simpler tasks. For instance, when using Llama3-8B for graph reading tasks, such as determining node count, edge count, and calculating the maximum, minimum, and average node degrees, Table~\ref{readgraph} reveals that, except for node count and maximum degree in smaller graphs, other graph properties are incorrectly identified. In large-scale graphs, LLM performance significantly declines. Therefore, for more complex tasks, such as finding optimal paths or solving combinatorial problems, LLMs struggle to deliver accurate solutions.

    % \item \textbf{Computational cost:} The computational cost of using LLMs for graph reasoning tasks is prohibitively high in most cases. These models are resource-intensive, requiring significant memory and computational power for inference, which poses a major bottleneck for real-world applications where efficiency and scalability are critical. This issue becomes even more severe for tasks involving large graphs, where encoding the graph structure alone would consume a huge sizeof tokens, or for tasks requiring repeated queries, as multiple inference calls can quickly become unsustainable.
    % In contrast, traditional graph algorithms are highly optimized for such tasks, often providing exact solutions with minimal computational overhead. This stark difference in efficiency highlights a key limitation of LLMs, making them impractical for widespread deployment in scenarios demanding low latency and high scalability.

    \item \textbf{Prohibitive Computational Cost:} %Using LLMs for graph computational tasks is costly. These models are resource-intensive, requiring significant memory and computational power for inference. Furthermore, the entire graph structure needs to be serialized and incorporated into prompt for LLM reasoning. For large graphs, it results in excessively long prompts, which significantly diminishes inference efficiency of LLMs. Table~\ref{time} shows even for polynomial-time problems like shortest path, LLMs still take about 20s to solve a query in a graph with only $10^2$ nodes. In contrast, traditional graph algorithms are highly optimized, providing exact solutions with minimal overhead. For example, Dijkstra algorithm can easily handle queries of $10^6$ nodes within 1s. This stark difference in efficiency highlights a key limitation of LLMs, making them impractical for widespread deployment in scenarios demanding low latency and high scalability.
    Leveraging LMs for graph computational tasks incurs significant costs. These models are highly resource-intensive, requiring substantial memory and computational power for inference. Furthermore, the entire graph structure needs to be serialized and incorporated into the prompt for LLM reasoning, which becomes particularly problematic for large graphs. This results in excessively long prompts that severely impact inference efficiency. As shown in Figure~\ref{runtime}, even for polynomial-time problems such as shortest path computation, LLMs require around 20 seconds to process a query on a graph with just 100 nodes. In contrast, traditional graph algorithms provide exact solutions with minimal computational overhead. For instance, Dijkstra~\cite{dijkstra} algorithm can efficiently handle queries involving up to $10^6$ nodes within one second. This significant disparity in efficiency underscores a major limitation of LLMs, rendering them impractical for use in applications that require low latency and high scalability.

\end{enumerate}

When humans solve graph computational problems, they typically follow three steps: problem understanding, algorithm design, and algorithm execution. Initially, they focus on comprehending the problem's definition and analyzing the structure of the input graph. Next, they gather relevant information and design an appropriate algorithm. Finally, they choose useful tools to execute the designed algorithm. In contrast, LLMs excel in the second step due to their extensive knowledge and ability to process and reason over unstructured text. However, LLMs face significant limitations in the other two steps. When understanding tasks,  LLMs struggle to grasp fundamental elements of graphs, as they are primarily designed to handle unstructured text rather than highly structured graph data. Furthermore, LLMs are unable to execute algorithms with precision, as hallucinations often cause them to overlook critical details in algorithms.

% In this work, we aim to propose \red{[Name]}, a novel framework that decomposes graph problem-solving into two stages: (1) leveraging LLMs for \textbf{problem understanding and algorithm design}, and (2) using generated code to \textbf{algorithm execution}.
% By allowing LLMs to focus on their strengths in reasoning and code generation, while delegating execution to more efficient and reliable methods, our approach not only reduces the computational cost of LLM usage but also significantly improves performance and accuracy in solving graph-related tasks.

%Inspired by human's approach to solve graph tasks, we propose a novel framework called \textbf{PIE} (\underline{\textbf{P}}seudocode-\underline{\textbf{I}}njection-\underline{\textbf{E}}nhanced LLM Reasoning on Graph Computational Tasks), which decomposes graph problem-solving into three steps: (1) allocating the understanding of task and graph structure to LLMs and interpreter respectively, (2) leveraging LLMs to design algorithm by writing code, and (3) using generated code for algorithm execution. By allowing LLMs to focus on their strengths in reasoning and code generation, while delegating execution to more efficient and reliable methods, our approach not only reduces the computational cost of LLM usage but also significantly improves accuracy in various graph computational tasks.

Inspired by the human approach to solving graph tasks, we introduce a novel framework called \textbf{PIE} (\underline{\textbf{P}}seudocode-\underline{\textbf{I}}njection-\underline{\textbf{E}}nhanced LLM Reasoning for Graph Computational Tasks). This framework decomposes the graph problem-solving process into three key steps: (1) %delegating the task understanding and graph structure analysis to LLMs and an interpreter, respectively, 
allocating the understanding of task and graph structure to LLMs and an interpreter, respectively
(2) leveraging LLMs to design the algorithm by writing code, and (3)  using generated code for algorithm execution. By allowing LLMs to focus on their strengths in reasoning and code generation, while delegating the execution to more efficient and reliable methods, our approach reduces the computational cost of LLM usage and significantly enhances accuracy across various graph computational tasks.
%
% For complex NP-hard problems, LLMs often struggle to provide suitable algorithms due to outdated knowledge and, in experiments, frequently resort to brute-force solutions. To address this issue and achieve better results, we propose a \textit{pseudocode injection} method, which enhances LLMs' understanding and improves their graph reasoning capabilities. Additionally, this approach allows LLMs to incorporate insights from the latest research, enabling them to generate more efficient and effective code.
%
%In experiments, we find that LLMs struggle to output efficient algorithms other than brute-force solutions for NP-complete tasks. To address this issue, 
In particular, we propose a \textbf{\textit{pseudocode injection}} method to enhance LLMs in code generation. %This improvement also helps LLMs to incorporate insights from the latest research and generate more efficient code. Meanwhile, to reduce the probability of errors during code executation, we perform multiple trials-and-errors on simple test samples and instruct LLMs to automatically correct errors. Finally, we choose the best-performing code as LLM's output and input into interpreter for further evaluation. The experimental results on nine graph reasoning tasks demonstrate that PIE significantly improves accuracy and substantially reduces the inference cost of LLMs. 
 This approach allows LLMs to integrate insights from the latest research, enabling the generation of more efficient code. To minimize execution errors, we conduct multiple trials using simple test samples and instruct LLMs to automatically correct any errors. The most effective code is then selected as the final output from the LLM and input into an interpreter for further evaluation. Experimental results across nine graph reasoning tasks demonstrate that PIE significantly improves accuracy while considerably reducing the inference cost of LLMs.

Our contributions are summarized as follows:
\begin{enumerate}[(1)]
    \item We propose a novel framework PIE to augment LLMs' ability to solve graph computational tasks. %, where LLMs understand task description and generate code, and interpreter executes the code to get final results.
    In this framework, LLMs understand the task description and generate the corresponding code, while an interpreter executes the code to produce the final results.
    
    \item To help LLMs generate efficient code, we propose pseudocode injection method and use multiple trial-and-error to ensure the output code runs correctly.
    \item Extensive experiments show that PIE achieves significantly higher accuracy than other methods, %with lowest cost.
     while maintaining the lowest computational cost.
\end{enumerate}

\begin{figure*}[htp]
\vspace{-0.1in}
\centerline{\includegraphics[width=0.8\linewidth]{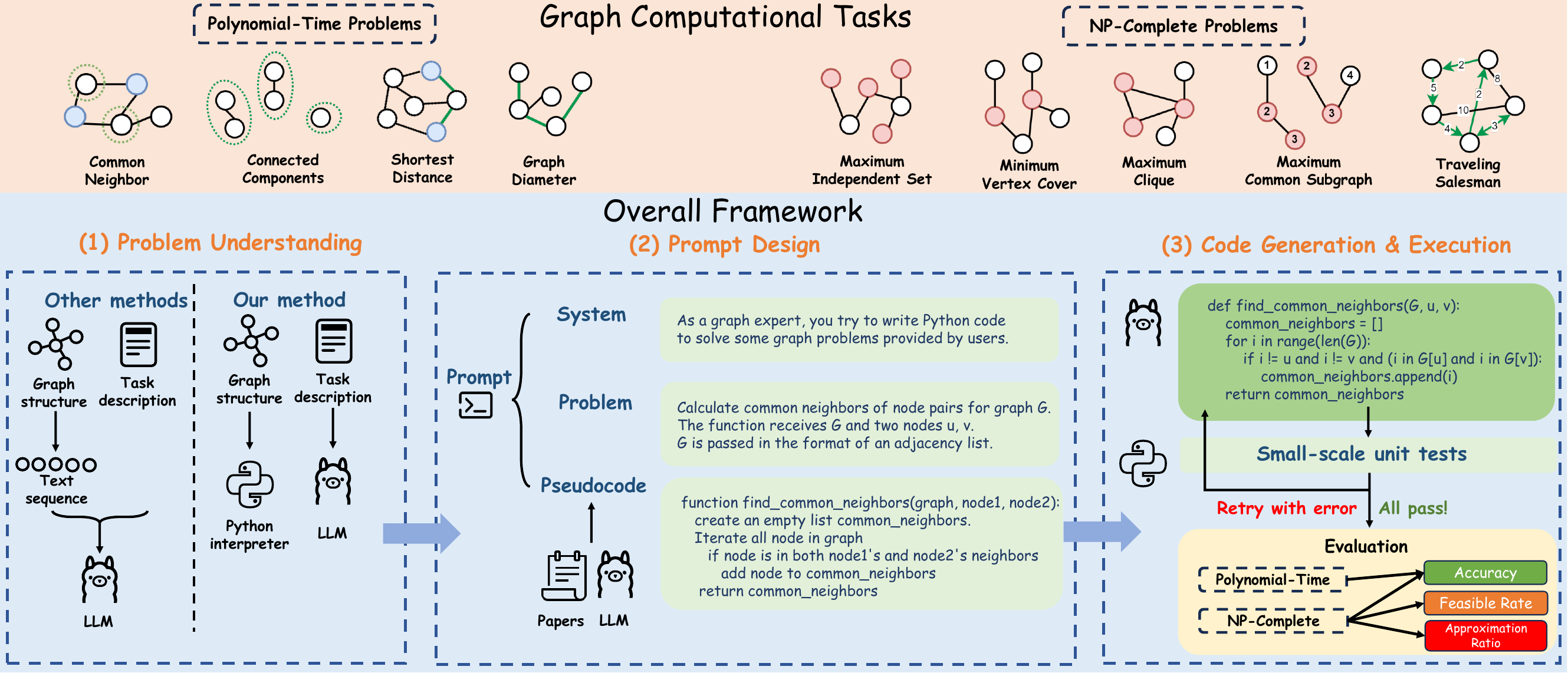}}
\vspace{-0.1in}
\caption{Nine graph tasks and the overview of our framework. In ``Problem Understanding'' phase, our approach allocates the understanding of task description and graph structure to LLM and Python interpreter respectively, which reduces the probability of errors arising from graph reading. In phase ``Prompt Design'', we design three types of prompts, including pseudocode extracted from papers, to guide the LLM to generate efficient code. Finally, in ``Code Generation and Execution'' phase, we utilize a small-scale dataset and perform multiple trials-and-errors to ensure LLM's output code to run correctly, which is then employed on other large-scale graphs.}
\label{framework}
\vspace{-0.15in}
\end{figure*}

\begin{table}[t]
\small
\centering
\setlength{\tabcolsep}{0.5mm}
\begin{tabular}{ccccccc}
\toprule
\multirow{2}{*}{\begin{tabular}[c]{@{}c@{}}Graph Node \\Number $|N|$\end{tabular}} & \multirow{2}{*}{\begin{tabular}[c]{@{}c@{}}Graph\\ Prompt\end{tabular}} & \multirow{2}{*}{\begin{tabular}[c]{@{}c@{}}Node\\ Number\end{tabular}} & \multirow{2}{*}{\begin{tabular}[c]{@{}c@{}}Edge\\ Number\end{tabular}} & \multirow{2}{*}{\begin{tabular}[c]{@{}c@{}}Max \\ Degree\end{tabular}} & \multirow{2}{*}{\begin{tabular}[c]{@{}c@{}}Min \\ Degree\end{tabular}} & \multirow{2}{*}{\begin{tabular}[c]{@{}c@{}}Avg \\ Degree\end{tabular}} \\
 &         &            &           &                  &          &    \\ \midrule
\multirow{3}{*}{\begin{tabular}[c]{@{}c@{}}Small\\ $[3,10]$\end{tabular}}       & Edge List           & 85.7            & 71.4         & 76.5           & 12.2         & 3.1                                                                    \\
   & Adj     & 87.0          & 25.0          & 75.0          & 36.0                 & 14.0                                                                   \\
    & Adj NL          & 81.0              & 22.0           & 65.0       & 57.0             & 25.0                                                                   \\ \midrule
\multirow{3}{*}{\begin{tabular}[c]{@{}c@{}}Large\\ $[20,100]$\end{tabular}}          & Edge List                                                               & 4.0                   & 0.0                   & 2.0                   & 0.0                                                                    & 0.0     \\
  & Adj       & 13.0     & 0.0         & 2.0        & 0.0      & 0.0          \\
     & Adj NL          & 9.1            & 0.0    & 2.0          & 0.0           & 0.0      \\ \bottomrule
\end{tabular}
\vspace{-0.1in}
\caption{Accuracy(\%) of performing simple graph-reading tasks with different forms of graph prompts. ``Adj'' means adjacency table, while ``Adj NL'' denotes adjacency table in natural language. Please refer to Appendix~\ref{example_graph_structure} for specific examples.}
\label{readgraph}
\vspace{-0.1in}
\end{table}

\section{Graph Computational Tasks}
\label{taskintro}
%Graph computational tasks involve focus on mining relationships between elements on the graph.
% Graph computational tasks involve mining relationships between nodes on the graph.
% We select nine challenging tasks, which are divided into two groups: polynomial-time and NP-complete. Figure~\ref{framework} shows examples of these tasks.
% Graph computational tasks involve extracting relationships between nodes within a graph. 
In this paper, we investigate nine representative and challenging graph computational tasks, categorized into two groups: polynomial-time problems and NP-complete problems. %Examples of these tasks are shown in Figure~\ref{framework}.
Figure~\ref{framework} presents examples of these tasks.
\\
\noindent\textbf{{Polynomial-Time Tasks}}

\begin{enumerate}[(1)]
    \setlength{\itemsep}{-0.5pt}
    % \item \textbf{Common Neighbors(CN)}: Given two nodes $u$ and $v$ in a graph $G$, identify the set of nodes that are mutual neighbors of both $u$ and $v$.
    \item \textbf{Common Neighbors (CN)}: Find common nodes connected with two given nodes $u$ and $v$ in a graph $G$.
    % \item \textbf{Connected Components}: In an undirected graph $G$, determine one or more sets of nodes such that each node within a set can be reached from any other node in the same set via direct or indirect edges.
        \item \textbf{Connected Components (CC)}: Find all disjoint connected components in an undirected graph $G$. Any two nodes in a CC can be connected by a path within this component.
    % \item \textbf{Shortest Path}: Given two nodes $u$ and $v$ in a graph $G$,  locate a path connecting $u$ and $v$, denoted as $(u, node1, node2, ... , v)$, ensuring this path is the shortest among all possible routes between $u$ and $v$.
        \item \textbf{Shortest Path (SP)}: Given two nodes $u$ and $v$ in an unweighted and undirected graph $G$, find the shortest path connecting the two nodes.
    % \item \textbf{Graph Diameter}: For an undirected graph $G$, the Graph diameter is defined as the longest shortest path between any two nodes in $G$. Essentially, it represents the greatest distance between any pair of nodes within the graph.

        \item \textbf{Graph Diameter (GD)}: For an undirected graph $G$, find the maximum distance between any node pair, where distance is the length of shortest path between two nodes.
\end{enumerate}
\noindent\textbf{{NP-Complete Tasks}}
\begin{enumerate}[(1)]
\setlength{\itemsep}{-0.5pt}
    % \item \textbf{Maximum Independent Set(MIS)}: In an undirected graph $G$, problem solver should identify a set of vertices with the largest cardinality such that no two vertices in this set are adjacent to each other.
    \item \textbf{Maximum Independent Set (MIS)}: Find the largest node set in an undirected graph $G$ such that no two nodes are adjacent.
    
    % \item \textbf{Minimum Vertex Cover(MVC)}: Given an undirected graph $G$, determine a vertex set $S\subset V$ such that every edge in G is incident to at least one vertex in $S$, and the cardinality of $S$ is minimized.

        \item \textbf{Minimum Vertex Cover (MVC)}: Find the smallest node set $U$ in an undirected graph $G(V,E)$ such that for any edge $e=(u,v)\in E$, $u\in U$ or $v\in U$.
        
    % \item \textbf{Maximum Clique Problem(MCP)}: Given an undirected graph $G$, the task is to find a complete subgraph with the largest number of vertices. In a complete subgraph, there exists an edge between every pair of vertices.

        \item \textbf{Maximum Clique Problem (MCP)}: Find the largest complete subgraph in an undirected graph $G$, where any two nodes are neighbors.
    
    % \item \textbf{Maximum Common Subgraph(MCS)}: The solution of this task is to find the largest subgraph with the same structure in two or more graphs. This subgraph must not only contain the maximum number of nodes and edges but also be connected and isomorphic in the two graphs. In this article, we focus on the problem of finding the maximum common subgraph in two graphs.

    \item \textbf{Maximum Common Subgraph (MCS)}: Find the largest isomorphic induced subgraph between two undirected graphs, where isomorphism means that one subgraph can be transformed into the other by one-to-one mapping.

    % \item \textbf{Traveling Salesman Problem(TSP)}: In a given weighted graph $G$, the task is finding a Hamiltonian cycle that satisfies the following conditions: (1) the cycle visits each node in the graph exactly once and returns to the starting point, and (2) among all such cycles, the total distance is minimized.

        \item \textbf{Traveling Salesman Problem (TSP)}: Find the shortest Hamiltonian cycle in a weighted complete graph $G$, which %means traversing all nodes exactly once and returning to the starting node.
        visits each node exactly once and returns to the starting node.
        
\end{enumerate}

\section{Methodology}
\subsection{Overview of Framework PIE}
To mitigate LLM's incorrect understanding of graph structure and extremely long inference time, we propose a novel framework \textbf{PIE} (\underline{\textbf{P}}seudocode-\underline{\textbf{I}}njection-\underline{\textbf{E}}nhanced LLM Reasoning on Graph Computational Tasks) that decomposes graph reasoning into three steps: \textbf{problem understanding}, \textbf{prompt design}, and \textbf{code generation and executation}. The overview of PIE is illustrated in Figure~\ref{framework}. First, we instruct LLMs to focus on understanding the task description instead of the graph structure, leaving the latter to the interpreter. Then, we carefully design prompts that integrate algorithm-related pseudocode, which helps LLMs to generate efficient code. Finally, we ensure the correctness of LLM's output code through low-cost iterative refinements and apply it to other large-scale datasets.

\subsection{Problem Understanding}
%Previous works focus on the impact of different forms of graph serialization on reasoning results. However, how to serialize graphs might not be the most important since LLMs cannot understand graph structure, which is evident from their inability to correctly identify basic elements on graphs. Nevertheless, LLMs possess a fundamental understanding of various graph reasoning tasks. For example, they can point out the definition of the shortest path between two nodes. Thus, establishing a connection between LLMs, which contain background knowledge, and native graph structures becomes a key point.
Previous research has primarily focused on examining how different forms of graph serialization affect reasoning results. However, how to serialize graphs may not be the most critical factor, as LLMs struggle to comprehend graph structures, as evidenced by their inability to accurately identify basic graph elements. Nonetheless, LLMs possess a solid understanding of various graph reasoning tasks; for instance, they can identify the shortest path between two nodes. Therefore, the key challenge is to establish a connection between the background knowledge embedded in LLMs and the inherent structure of graphs.

%We adopt a completely new approach: the understanding of tasks and graph structures is allocated to LLMs and interpreters, respectively. The former generate task-related code, while the latter execute the code to get final results. 

To address this, we propose a new approach: task understanding and graph structure interpretation are delegated to LLMs and interpreters, respectively. LLMs generate task-related code, while interpreters execute the code to produce the final results.
%
%In this way, LLMs and interpreters leverage their respective strengths. LLMs  utilize knowledge about tasks to write code without focusing on the specific form of graph structure (they can access graph elements through abstract APIs). Interpreters can precisely figure out basic elements of the graph by specifying graph structure in a particular format, and execute code to obtain definitive results. Moreover, after LLMs generate correct code, interpreters can execute the same code for the same category of tasks without repeatedly calling LLMs, which reduces inference cost of LLMs.
%
This approach allows LLMs and interpreters to leverage their respective strengths effectively. LLMs utilize their task-related knowledge to generate code without the need to understand the specific graph structure, accessing graph elements through abstract APIs. On the other hand, interpreters precisely identify the fundamental components of the graph by specifying its structure in a particular format, and execute the generated code to obtain definitive results. Furthermore, once LLMs produce correct code, interpreters can execute the code for subsequent tasks in the same category without repeatedly invoking the LLM, thereby reducing the overall inference cost.

\subsection{Prompt Design}
% LLMs possess extensive knowledge about graph computational tasks. We need to design sophisticated prompts to guide them to fully leverage this knowledge and generate correct and efficient code. The prompt consists of three parts: system prompt, problem prompt, and pseudocode prompt.

% LLMs have a vast understanding of graph computational tasks. To fully utilize this knowledge and generate accurate and efficient code, we must design well-structured prompts. The prompt is composed of three key components: the system prompt, the problem prompt, and the pseudocode prompt.
LLMs may possess a comprehensive knowledge of graph computational tasks. To effectively leverage this knowledge and produce accurate, efficient code, it is essential to craft well-structured prompts. These prompts consist of three main components: the system prompt, the problem prompt, and the pseudocode prompt.

\subsubsection{System Prompt} In this part, %we ask LLM to act as an expert in graph algorithms and try to solve problems by writing code. To facilitate further processing, we also force LLM to provide output in a specified format. This part is shared among all tasks.
%In this section, 
we instruct the LLM to act as an expert in graph algorithms and attempt to solve problems by writing code. To facilitate subsequent processing, we also specify a required output format. This structure is consistent across all tasks.

\subsubsection{Problem Prompt} %In this part, we briefly describe task and clarify the objective of LLM generation. Through experiments, we find that directly generating complete executable code without error is quite challenging for LLMs, and LLMs tend to customize certain data structures, which is inconvenient for batch processing by the interpreter. To reduce the difficulty of LLM generation and align the format of LLM's output and interpreter's input, we require LLMs to write functions with specified input and output formats.
In this part, we provide a concise description of the task and clearly define the objectives for the LLM's output. Through experiments, we have observed that generating fully executable and error-free code is a significant challenge for LLMs. LLMs often create custom data structures, which complicates batch processing by the interpreter. To mitigate this challenge and ensure compatibility between the LLM's output and the interpreter's input, we instruct the LLM to write functions with specified input and output formats.

\subsubsection{Pseudocode Prompt}
%Using the above two prompts, LLMs can output task-related code. However, they tend to provide brute-force algorithms. For example, when processing TSP problems, LLMs would traverse all feasible solutions to find the shortest cycle, which runs slowly and cannot scale to large graph problems. Nevertheless, by directly inquiring whether there exists improved algorithm for TSP, LLMs can provide an affirmative response and describe the core idea of a greedy algorithm using natural language. This inspires us to employ explicit methods to guide LLMs to uncover their implicit information.
%
Using the prompts described above, LLMs can generate task-specific code. However, they often default to brute-force algorithms. For example, when solving the TSP problems, LLMs may explore all possible solutions to find the shortest cycle, leading to slow performance and limited scalability for large graph instances. However, directly asking whether a more efficient algorithm exists for TSP can prompt the LLM to acknowledge the possibility and describe the core idea behind a greedy algorithm in natural language. This insight motivates us to employ explicit methods to guide LLMs in uncovering their implicit knowledge.

%We collect pseudocode for various graph tasks, enabling the LLM to integrate the relevant knowledge and generate executable code. Specifically, for each graph task, we extract pseudocodes from multiple relevant papers and input them into an LLM to standardize their format. Then we instruct the LLM to analyze the time and space complexity as well as coding difficulty of different pseudocodes and select the optimal one to serve as the final pseudocode prompt.
%
To enhance the LLM’s capabilities, we collect pseudocode from various graph-related tasks, allowing the model to integrate relevant knowledge and generate executable code. Specifically, for each task, we extract pseudocode from multiple research papers, standardize their format, and input them into the LLM. We then instruct the model to analyze the time and space complexity, as well as the coding difficulty, of different pseudocodes and select the optimal one as the final prompt for generating code.

%PSEUDO~\cite{pse} also uses pseudocode to enhance reasoning capabilities of LLMs. However, our method of extracting pseudocode is more reasonable. Firstly, the optimal pseudocode is selected from multiple papers, reducing the probability of extracting brute-force algorithm. Secondly, we use the LLM to standardize pseudocode's format, lowering the difficulty of generating corresponding code and reducing hallucination phenomena. Moreover, PSEUDO inputs pseudocode and serialized graph structure as prompt into the LLM, which fails to help LLM to read graphs and significantly increases inference cost. In contrast, our method instructs LLMs to write code based on provided pseudocode and the interpreter ensures high accuracy of problem-solving.

PSEUDO \cite{pse} also utilizes pseudocode to enhance the reasoning capabilities of LLMs. However, our approach to extracting pseudocode is more powerful. First, we select the optimal pseudocode from multiple papers, reducing the probability of extracting brute-force algorithms. Second, we standardize the pseudocode format using the LLM, which simplifies the code generation process and reduces the occurrence of hallucinations. Moreover, PSEUDO combines pseudocode with serialized graph structures as input prompts for the LLM, which not only fails to assist the model in effectively reading graphs but also significantly increases inference costs. In contrast, our method instructs the LLM to generate code based on the provided pseudocode, while an interpreter ensures high accuracy in problem-solving.

\subsection{Code Generation and Execution}

%By combining three prompts mentioned above, LLMs can generate corresponding code. However, when the interpreter executes it, errors are inevitably encountered, such as syntax errors like missing key modules, or runtime errors like lacking attributes for an object. These errors are unrelated to the core idea and can be easily corrected by LLM itself. Thus, we propose the following two strategies to urge LLMs to review and output correctly executed code:
By integrating the three prompts mentioned above, LLMs can generate the corresponding code. However, when the code is executed by the interpreter, errors such as syntax issues (e.g., missing modules) or runtime problems (e.g., missing object attributes) may arise. These errors are typically unrelated to the core logic and can be easily addressed by the LLM itself. Thus, we propose the following two strategies to guide LLMs in reviewing and producing error-free code:

\begin{enumerate}[(1)]
    \item For each task, we generate 10 small-scale graphs $d_{small}$ (node number $|N|<10$). %For code $c$ output by LLM, a test function $f$ is used to evaluate its performance on $d_{small}$.
    The code $c$ generated by the LLM is evaluated using a test function  $f$ on $d_{small}$. If the code passes all tests, it is selected as the final output. If it fails any test, the LLM is required to regenerate the code.
    %If it passes all tests, $c$ is selected as the final output code. Otherwise, LLM needs to generate code again.
    \item %When encountering an error, the test function $f$ extracts the error message and returns it to the LLM for code correction, until the code runs correctly.
    In the event of an error, the test function $f$ extracts the error message and provides it to the LLM for correction, repeating this process until the code executes correctly.
\end{enumerate}

% The whole process is illustrated in Algorithm~\ref{trial}. To prevent the high cost of repeatedly calling LLMs, we set an upper limit $K$ on the number of retries. If the LLM cannot pass $d_{small}$ within $K$ trials, we select the best-performing code as the final output (for polynomial-time and NP-complete tasks, we use accuracy and approximation ratio as evaluation metrics). Notice that the above trial-and-error method is executed automatically without human intervention, thereby reducing the effect of human prior knowledge on the final result.

The whole process is outlined in Algorithm~\ref{trial}. To mitigate the high cost associated with repeatedly invoking LLMs, we impose an upper limit $K$ on the number of retry attempts. If the code fails to pass all tests on $d_{small}$ within $K$ trials, we select the best-performing code as the final output. For polynomial-time and NP-complete tasks, we use accuracy and approximation ratio as evaluation metrics. Notice that this trial-and-error process is executed automatically without human intervention, thereby reducing the effect of human bias on the final result.

\begin{algorithm}[t]
\caption{Code Generation and Execution}
	\label{trial}
    % \SetKwRepeat{Do}{do}{while}
	\KwIn{Combined prompt $p$, LLM $m$, Test function $f$, Trial upper limit $K$, small-scale graphs $d_{small}$.}
    \KwOut{Best-performing code $c_{best}$.}
	% \BlankLine
    $i=0;c_{tot}=\varnothing$;\\
    \While{$i<K$} {
        $c=m(p)$; $e=f(c,d_{small})$;\\
        \While{$e$ contains ``error''} {
            $c=m(p, e)$; $e=f(c,d_{small});$
        }

        \eIf {$e$ contains ``all pass''} {
            \Return $c$.
        } {
            $c_{tot} = c_{tot} \cup \{c\};$\\
        }
        $i = i+1;$\\
    }
    \Return best-performing code $c_{best}$ from $c_{tot}$.
\end{algorithm}

% \usepackage{multirow}
\begin{table*}[ht]
\small
\centering
\setlength{\tabcolsep}{1mm}
\begin{tabular}{ccccccccccc}
\toprule

                            &                          & \multicolumn{4}{c}{Polynomial-time Tasks}                                                                                       & \multicolumn{5}{c}{NP-complete Tasks}        \\ \specialrule{0em}{1.0pt}{1.0pt} \cline{3-11}
                               \specialrule{0em}{1.0pt}{1.0pt}
\multirow{-2}{*}{LLM}       & \multirow{-2}{*}{Method} & CN                           & CC                           & SD                           & GD                           & MIS   & MVC   & MCP    & MCS   & TSP   \\
\midrule
                            & GraphArena               & \small{72.4/68.6} & \small{37.1/37.1} & \small{30.5/16.2} & \small{29.5/12.4} & \small{33.3/8.6}  & \small{27.6/19.1}  & \small{35.2/16.2}   & \small{36.2/20.0}  & \small{1.9/0.0}  \\
                            & PSEUDO                   
                            & \small{54.3/24.8} & \small{57.1/25.6} & \small{58.1/34.3} & \small{33.3/17.1}                        & \small{55.2/2.9} & \small{28.6/13.3} & \small{40.0/15.2}  & \small{30.5/2.9} & \small{1.9/0.0}  \\
                            & Talk-like-a-graph        & \small{82.8/68.0} & \small{65.7/25.9} & \small{67.5/26.7}                        & \small{41.3/18.3}                        & \small{39.1/3.8} & \small{23.8/23.8} & \small{41.9/11.4}  & \small{17.1/4.8} & \small{1.9/0.0}  \\
                            & GraphInstruct            & \small{68.6/72.4} & \small{65.7/22.9}                        & \small{55.2/42.9}                        & \small{34.3/13.3}                        & \small{30.5/1.9} & \small{25.7/14.3} & \small{43.8/17.1}  & \small{31.4/2.9} & \small{1.9/0.0}  \\
                            & NLGraph                  & \small{33.3/25.7}                        & \small{56.2/21.9}                        & \small{41.9/23.8}                        & \small{32.4/19.1}                        & \small{46.7/2.9} & \small{33.3/17.1} & \small{44.8/12.4}  & \small{31.4/0.0} & \small{0.0/0.0}  \\ \specialrule{0em}{1.0pt}{1.0pt} \cline{2-11}
                               \specialrule{0em}{1.0pt}{1.0pt}
\multirow{-6}{*}{\begin{tabular}[c]{@{}l@{}}Llama3\\ \quad 8b\end{tabular}} & PIE (Ours)                     & \small{\textbf{100.0/100.0}}                       & \small{\textbf{100.0/100.0}}                       & \small{\textbf{100.0/100.0}}                       & \small{\textbf{100.0/100.0}}                       & \small{\textbf{98.0/87.6}} & \small{\textbf{98.8/95.2}} & \small{\textbf{100.0/100.0}} & \small{\textbf{85.0/28.7}} & \small{\textbf{80.2/27.4}} \\ \midrule 
                            & GraphArena               & \small{94.3/68.6} & \small{90.5/37.1} & \small{90.4/58.1} & \small{44.8/11.4} & \small{50.5/7.6}  & \small{21.9/10.5} & \small{63.8/16.2}  & \small{66.7/20.0}  & \small{0.0/0.0}  \\
                            & PSEUDO                   & \small{82.9/57.1} & \small{70.5/41.9} & \small{84.8/57.1} & \small{41.9/27.6}                       & \small{59.1/6.7} & \small{50.5/17.1} & \small{55.2/21.0} & \small{50.5/17.1} & \small{7.6/0.0}  \\
                            & Talk-like-a-graph        & \small{86.7/68.6} & \small{98.1/67.6} & \small{99.1/76.2}                        & \small{56.2/18.1}                        & \small{40.0/7.6} & \small{22.9/3.8} & \small{68.6/21.0} & \small{40.0/6.7} & \small{8.6/0.0} \\
                            & GraphInstruct            & \small{85.7/73.3} & \small{91.4/56.2}                        & \small{96.2/70.5}                        & \small{40.0/22.9}                     & \small{35.2/4.8} & \small{22.9/8.6} & \small{54.3/24.8}  & \small{54.3/6.7} & \small{5.7/0.0}  \\
                            & NLGraph                  & \small{73.3/41.0}                        & \small{81.9/33.3}                        & \small{83.8/49.5}                       & \small{47.6/24.8}                       & \small{41.0/2.9} & \small{38.1/14.3} & \small{52.4/8.6}  & \small{57.1/6.7} & \small{1.9/0.0} \\ \specialrule{0em}{1.0pt}{1.0pt} \cline{2-11}
                               \specialrule{0em}{1.0pt}{1.0pt}
\multirow{-6}{*}{\begin{tabular}[c]{@{}l@{}}Llama3\\ \quad 70b\end{tabular}} & PIE (Ours)                      & \small{\textbf{100.0/100.0}}                       & \small{\textbf{100.0/100.0}}                       & \small{\textbf{100.0/100.0}}                       & \small{\textbf{100.0/100.0}}                       & \small{\textbf{98.0/87.6}} & \small{\textbf{98.8/95.2}} & \small{\textbf{100.0/100.0}} & \small{\textbf{91.2/26.1}} & \small{\textbf{80.2/27.4}} \\ \bottomrule

\end{tabular}
\vspace{-0.1in}
\caption{Accuracy(\%) comparison among different methods on small/large graphs. Higher is better.}
\vspace{-0.1in}
\label{acc}
\end{table*}

\begin{figure*}[t]
\vspace{-0.15in}
    \centering
    \subfloat[Llama3-8b small graph]{\includegraphics[width=0.47\linewidth]{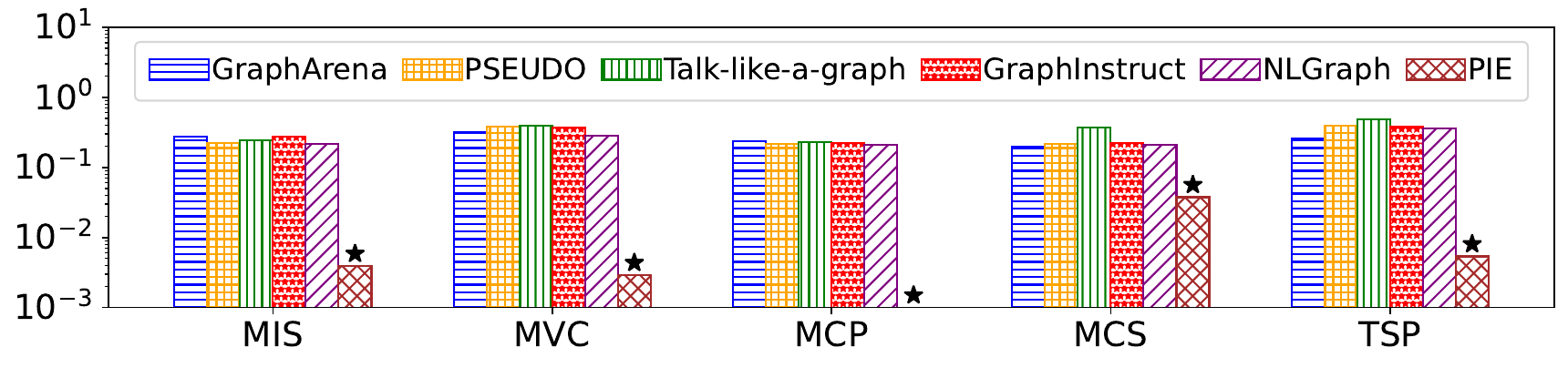}}%
    \label{ratio-8b-small}
    % \vspace{-0.1in}
    \subfloat[Llama3-8b large graph]{\includegraphics[width=0.47\linewidth]{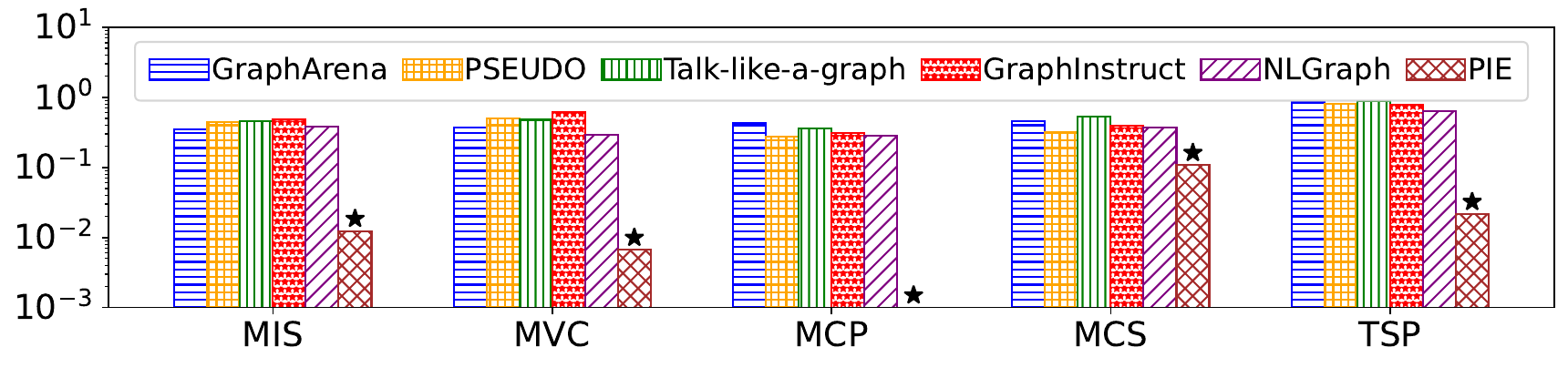}}%
    \label{ratio-8b-large}\\
        \vspace{-0.15in}
    \subfloat[Llama3-70b small graph]{\includegraphics[width=0.47\linewidth]{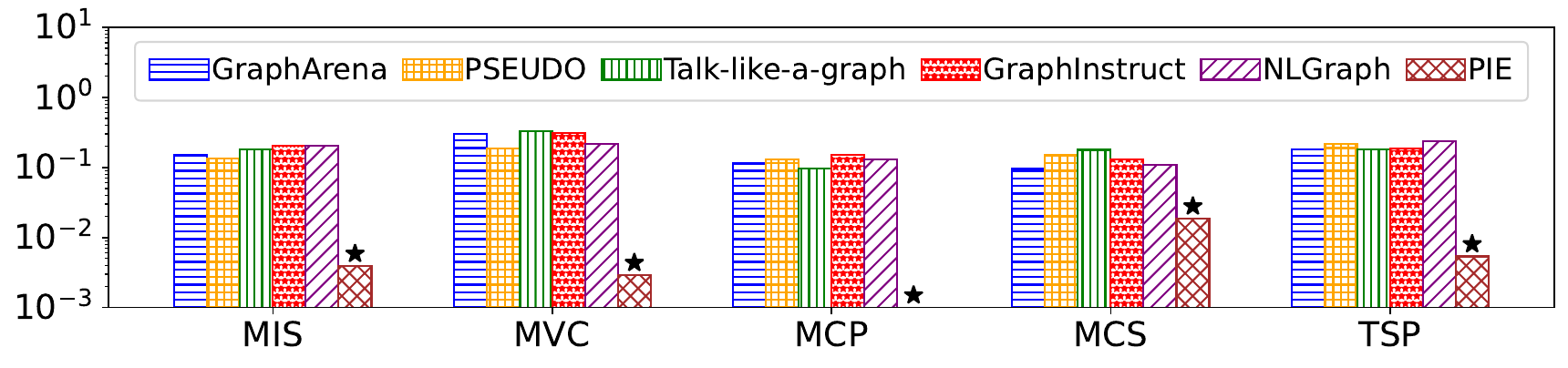}}%
    \label{ratio-70b-small}
        % \vspace{-0.1in}
    \subfloat[Llama3-70b large graph]{\includegraphics[width=0.47\linewidth]{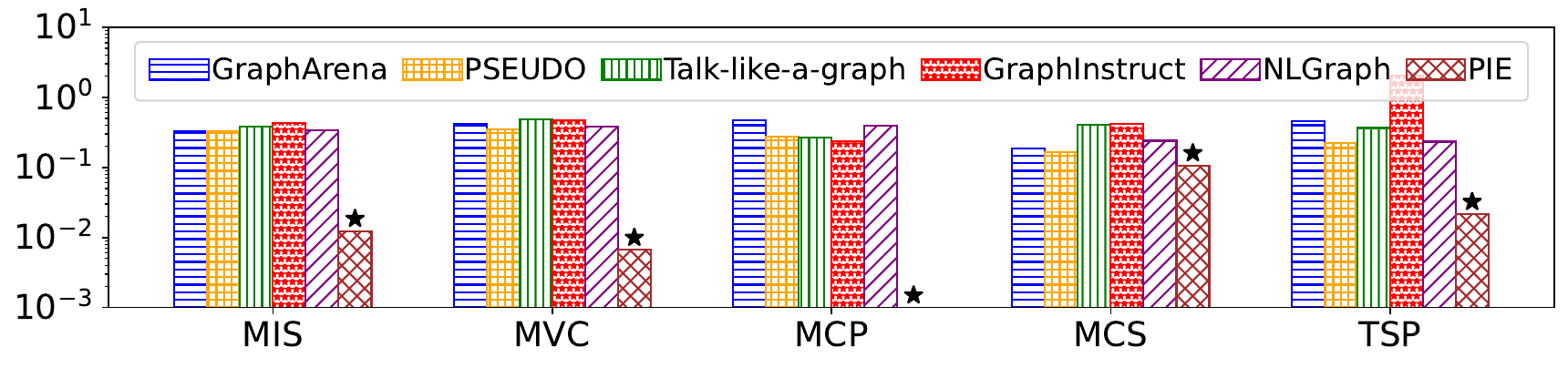}}%
    \label{ratio-70b-large}
    \vspace{-0.1in}
    \caption{Approximation ratio comparison of various methods on different NP-complete tasks. Lower is better.}
    \label{ar}
    \vspace{-0.2in}
\end{figure*}

\section{Experiments}
\subsection{Experimental Settings}
\noindent\textbf{Tasks and Datasets.} Nine graph computational tasks are introduced in Section~\ref{taskintro}. For each task, we follow GraphArena to construct 500 small- and large-scale test graphs. 

% Additionally, we generate two forms of graph structures: edge list format and natural language description format, to satisfy input requirements of different algorithms.

% \noindent\textbf{Tasks and Datasets.} Similar to GraphArena~\cite{grapharena}, we select nine classic graph reasoning tasks, including four polynomial-time (Common Neighbors (CN), Connected Components (CC), Shortest Distance (SD) and Graph Diameter (GD)) and five NP-complete ones (Maximum Independent Set (MIS), Minimum Vertex Cover (MVC), Maximum Clique (MC), Maximum Common induced Subgraph (MCS) and Traveling Salesman Problem (TSP)). For each task, we follow GraphArena to construct 500 small- and large-scale test graphs. Additionally, we  generate two forms of graph structures: edge list format and natural language description format, to satisfy input requirements of different algorithms.

% GraphArena is a tool proposed by xxx for evaluating large language models' ability to solve graph computation problems. This dataset contains 10 graph computation tasks, including 6 polynomial-time tasks and 4 NP-complete tasks. Each task is divided into two types of datasets: large graphs and small graphs, with 500 specific problems under each graph scale. In this paper, we utilize the method of extracting graph data employed by the GraphArena research team during the dataset construction process to build two evaluation datasets: one described in natural language form and the other represented using graph adjacency lists. This facilitates comparative experiments between different methods.

\noindent\textbf{Metrics.} 
% To comprehensively evaluate the performance of various algorithms, we consider two aspects of metrics:
We consider two aspects of metrics:
\begin{enumerate}[(1)]
    \item \textbf{Prediction precision}: %for polynomial-time tasks, we evaluate them based on Accuracy.
    For $T$ test graphs, suppose prediction and ground-truth value are $p_i$ and $p^{*}_i$, then $Accuracy=\frac{1}{T}{|\{i|p_i=p^{*}_i\}|}$. For NP-complete tasks, in addition to Accuracy, we include two metrics: $Feasible\;Rate (FR)$ and $Approximation\;Ratio =\frac{1}{T}\sum_{i}\frac{|p_i-p^{*}_i|}{|p^{*}_i|}$. For MIS, MCP, and MCS, ``FR'' represents the probability that prediction is no larger than the ground truth, whereas for MVC and TSP, it denotes the opposite scenario.
    \item \textbf{Efficiency and costs}: we also compare the costs of different algorithms, primarily measured by the cost of a single call to LLMs and the total number of calls.
\end{enumerate}
% \noindent\textbf{Metrics.} For four polynomial-time tasks, the evaluation is based on the accuracy of the answers (the proportion of correct answers). For the 4 NP-complete tasks, the evaluation is based on the overlap between the obtained answers and the precise solutions (see formula).

\noindent\textbf{Baselines.}
There have been many works aiming to enhance LLM's graph reasoning ability. We select five representative single-agent ones as our competitors: GraphArena~\cite{grapharena}, PSEUDO~\cite{pse}, Talk-like-a-graph~\cite{talklikeagraph}, GraphInstruct~\cite{graphinstruct}, and NLGraph~\cite{nlgraph}. The primary distinctions lie in how to serialize graph structure and utilize external knowledge (specific examples can be found in Appendix~\ref{prompt_example_by_others}). We do not compare with methods that finetunes LLMs such as GraphWiz~\cite{graphwiz} and GCoder~\cite{gcoder} since our approach does not update model's parameters. We also do not compare with multi-agent methods like GraphTeam~\cite{graphteam} because their LLM inference cost is significantly higher than single-agent ones.

\begin{figure*}[t]
\vspace{-0.2in}
    \centering
    \subfloat[Llama3-8b small graph]{\includegraphics[width=0.2\linewidth]{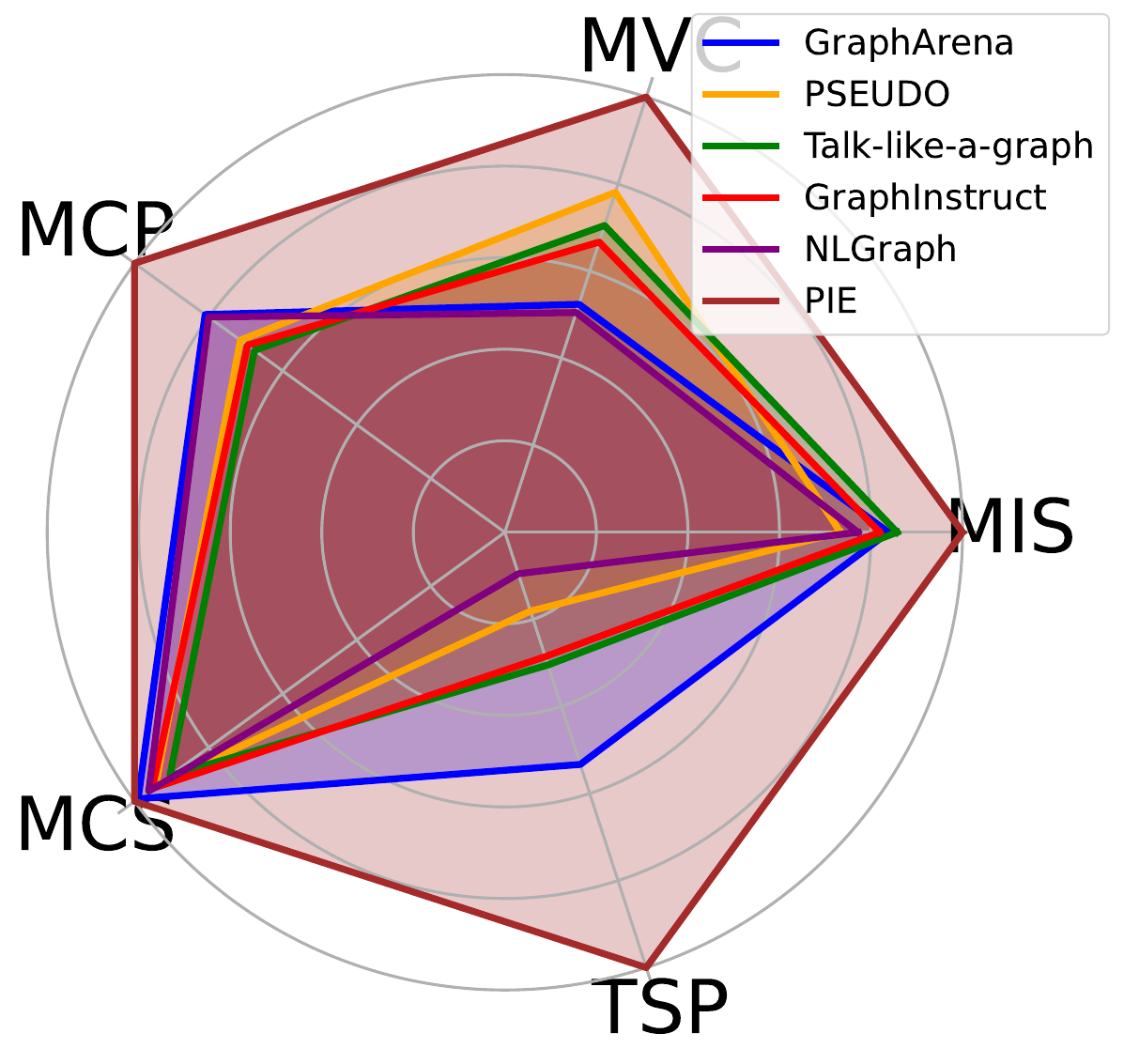}}%
    \label{feasible-8b-small}
    \subfloat[Llama3-8b large graph]{\includegraphics[width=0.2\linewidth]{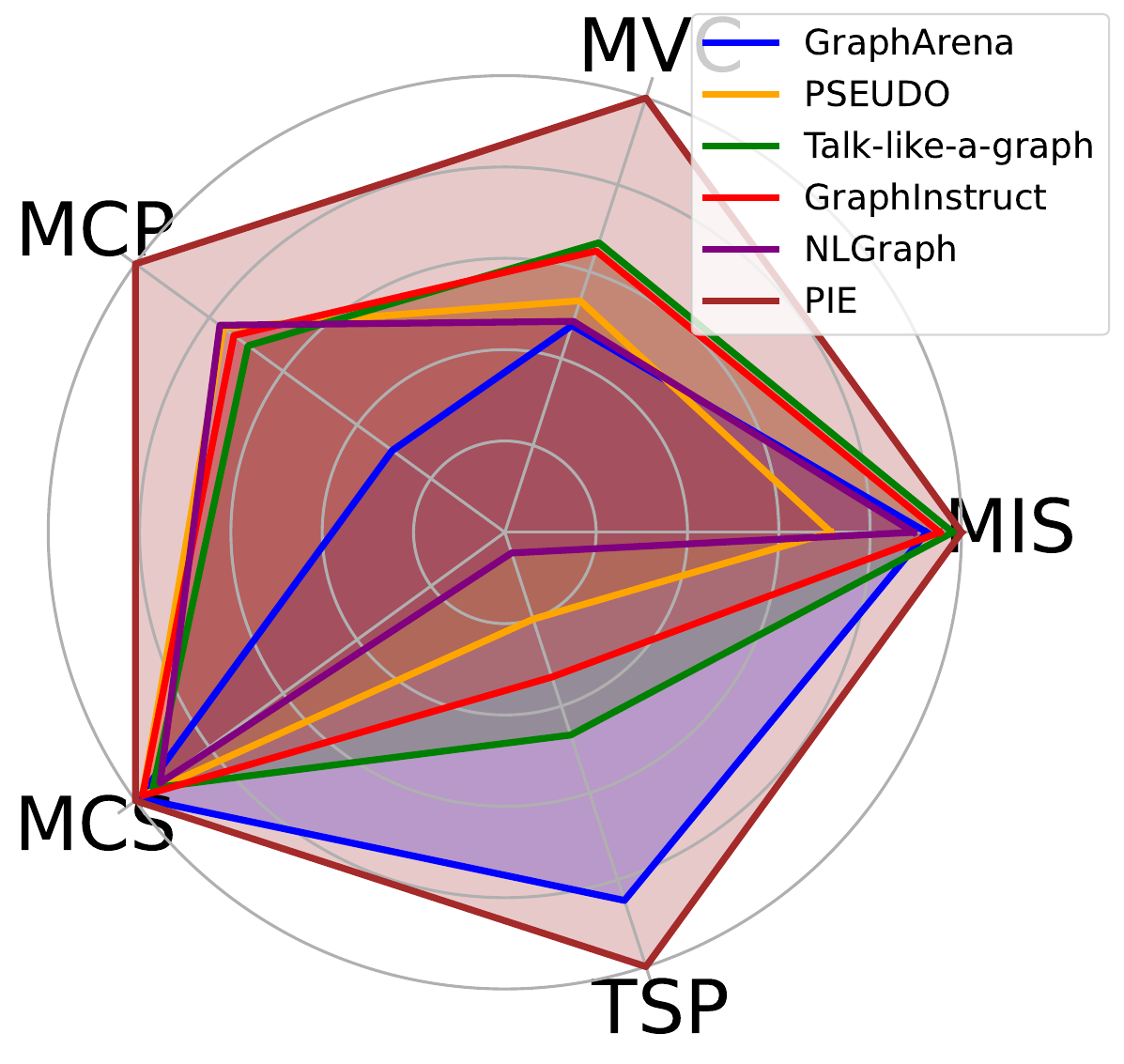}}%
    \label{feasible-8b-large}
    % \vspace{-0.1in}
    \subfloat[Llama3-70b small graph]{\includegraphics[width=0.2\linewidth]{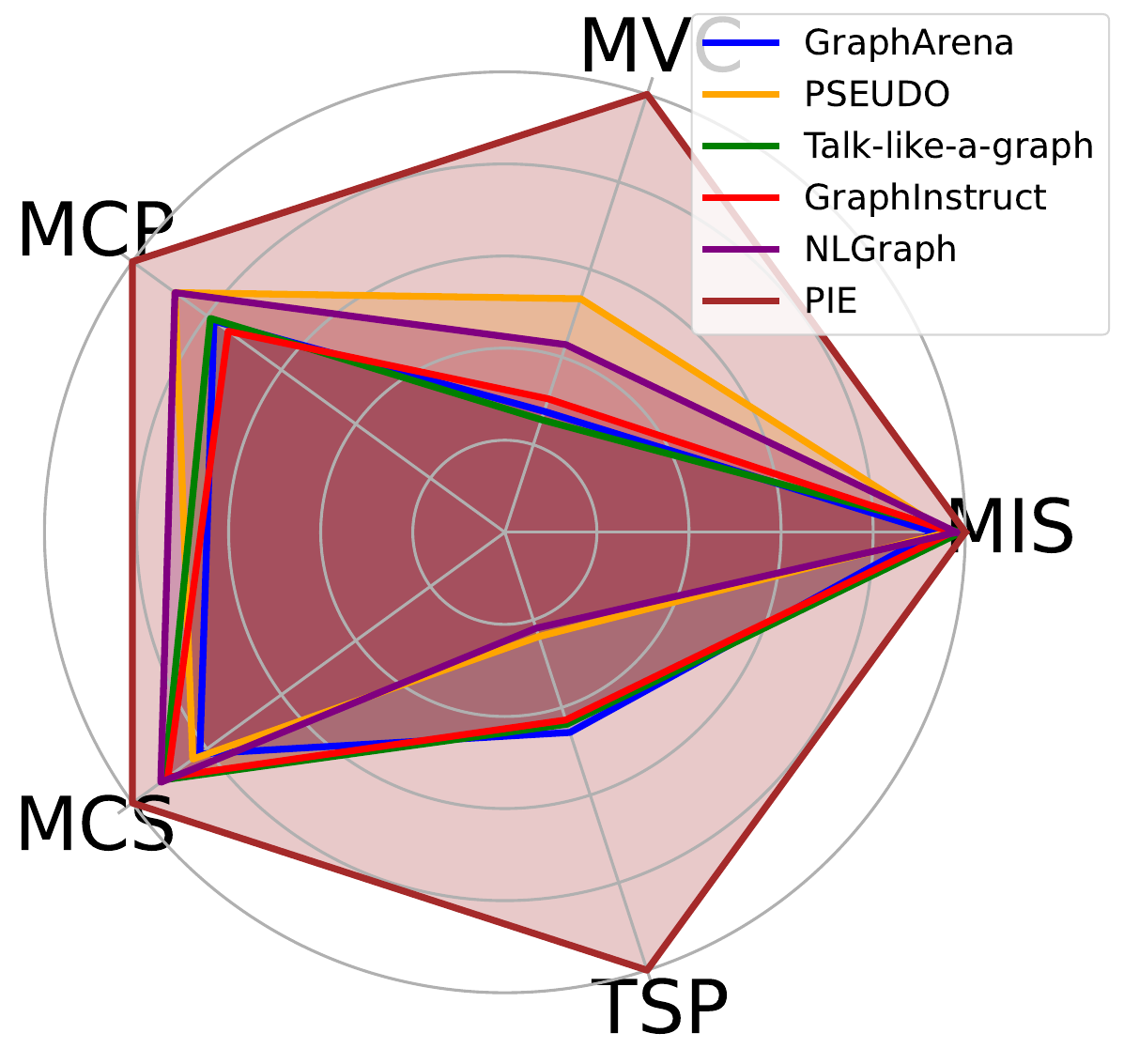}}%
    \label{feasible-70b-small}
    \subfloat[Llama3-70b large graph]{\includegraphics[width=0.2\linewidth]{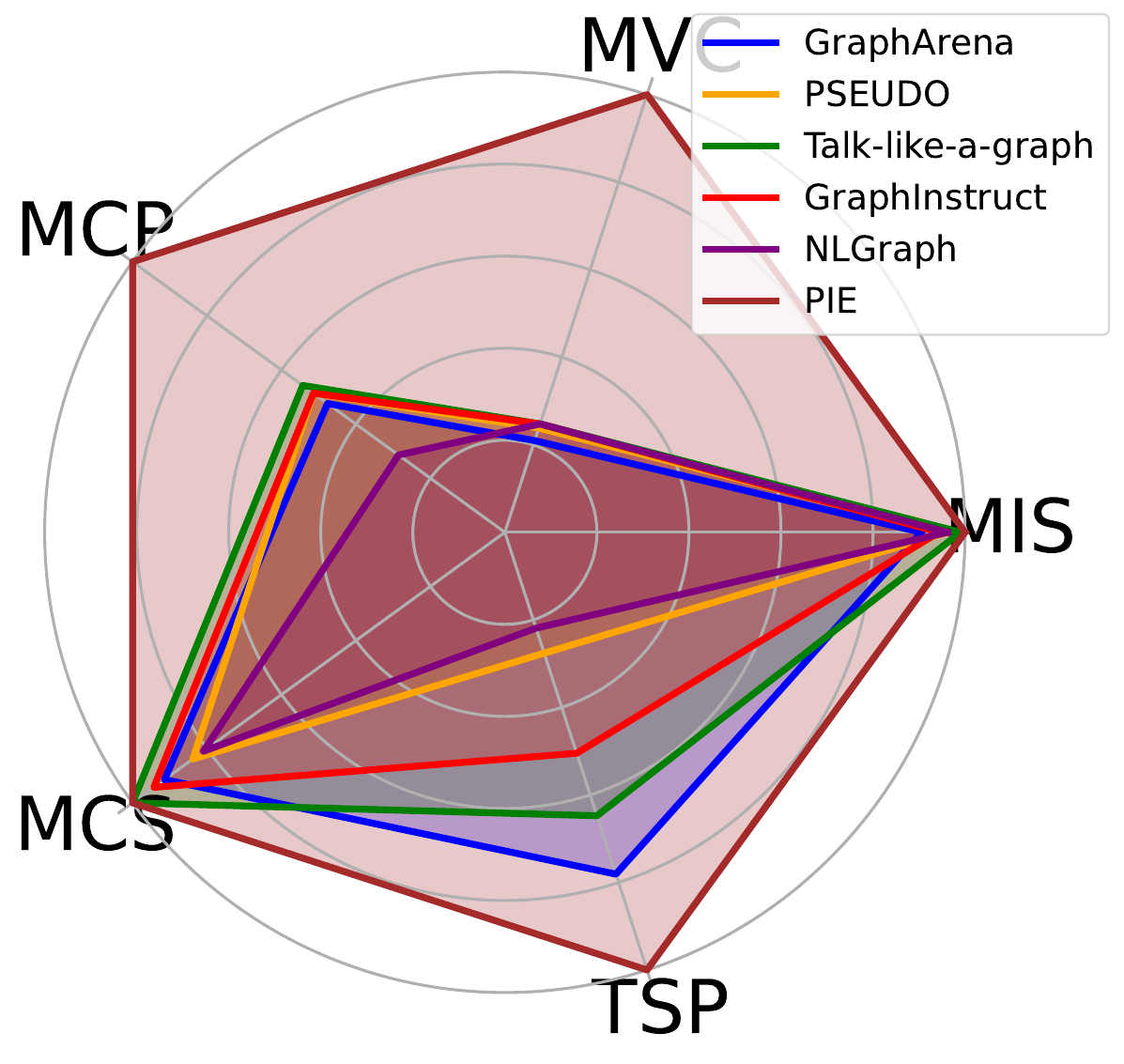}}%
    \label{feasible-70b-large}
    \vspace{-0.1in}
    \caption{Feasible rate comparison of various methods on different NP-complete tasks. Higher is better.}
    \label{fr}
    \vspace{-0.15in}
\end{figure*}

\noindent\textbf{Other Detailed Setups.}
We select three LLMs with different scales: Llama3-8b~\cite{llama3}, Llama3-70b, and DeepSeek-V3~\cite{deepseek-v3}. Due to limited space, we put the results of DeepSeek-V3 in Appendix~\ref{deepseek_exp}. All experiments run on a machine equipped with 8 NVIDIA A6000s. For PIE, we set the maximum number of retries for code generation per task to 10. For other methods, we follow the optimal settings as described in their papers.

\subsection{Performance Comparison}
\subsubsection{Prediction Precision}
Table~\ref{acc} compares the accuracy of various methods on different graph computational tasks. Overall, our method significantly outperforms other baselines. On polynomial-time tasks, our method achieves 100\% accuracy, and this result is independent of the LLM backbone and graph size. For NP-complete problems, our approach also ensures a high level of precision, especially on large graphs, with an average improvement of $60.3$ and $57.9$ percentage points compared to other methods with llama3-8b and llama3-70b backbone. 

The substantial enhancement is derived from circumventing graph reading by LLMs. Instead, they are applied to understand graph tasks from a broad perspective and focus on code generation that they excel in. The analysis of graph structure is delegated to interpreter, which can identify basic graph elements correctly. Moreover, the code generated by LLMs undergoes trial-and-error test to guarantee that it runs correctly, ensuring excellent performance on other datasets.

For NP-complete tasks, we further compare approximation ratios and feasible rates of different algorithms. As shown in Figure~\ref{ar}, the approximation ratio of our method is less than $\frac{1}{10}$ of other methods. Especially on MCP task, it equals to 0, indicating that the predictions are all correct. Figure~\ref{fr} illustrates the proportion of predictions that can be feasible solutions. Clearly, our approach achieves a perfect 100\% across all tasks. In contrast, other methods struggle to ensure even a feasible solution due to the lack of constraints on prediction. Especially for TSP task, whose ground-truth value's range can be extremely vast, it is quite challenging for them to control predictions into a reasonable range. Thanks to writing code by LLMs instead of using them to directly output an answer, our method ensures to acquire a feasible solution. Moreover, an extremely low approximation rate guarantees obtained feasible solution is close to ground truth, thereby enhancing the practicality of our method.

\begin{figure*}[ht]
\vspace{-0.1in}
    \centering
    \subfloat[Llama3-8b]{\includegraphics[width=0.49\linewidth]{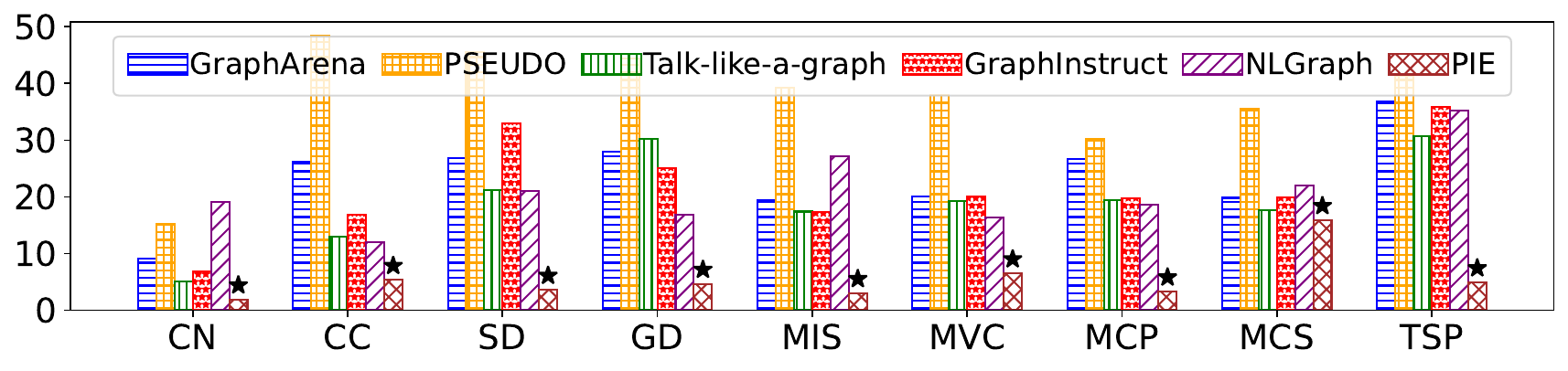}}%
    \label{llama8b-time}
    \subfloat[Llama3-70b]{\includegraphics[width=0.49\linewidth]{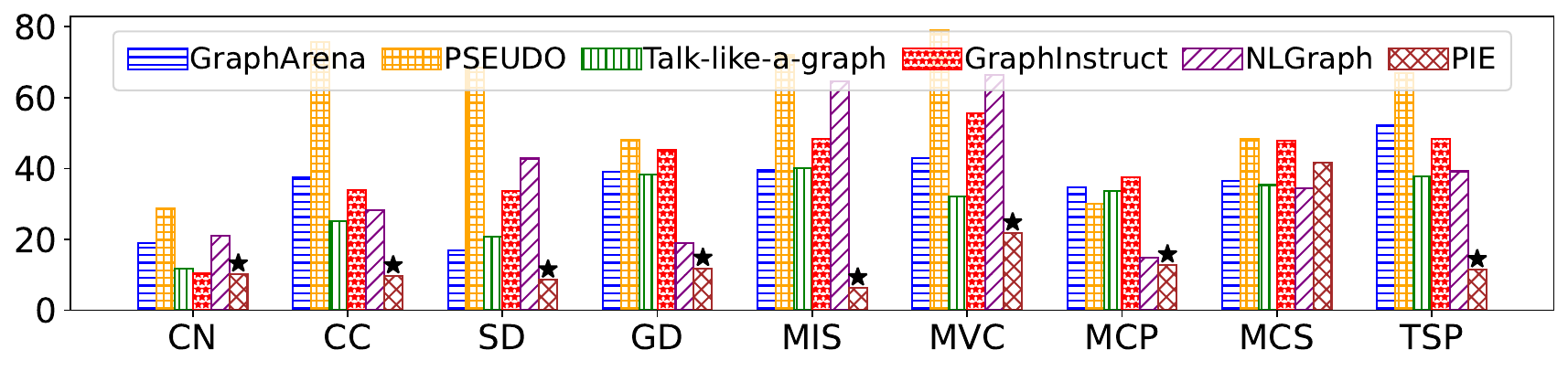}}%
    \label{llama70b-time}
    \vspace{-0.1in}
    \caption{Time cost comparison for a single call to LLMs, averaging small and large graphs.}
    \label{runtime}
    \vspace{-0.15in}
\end{figure*}

% \usepackage{multirow}
\begin{table}[t]
% \vspace{-0.1in}
\small
\centering
\setlength{\tabcolsep}{4.5mm}
\begin{tabular}{clc}
\toprule
LLM                         & Methods          & \# LLM Calls per Task \\
\midrule
\multirow{2}{*}{Llama3-8b}  & Others & 1000                               \\ 
                            & PIE (Ours)             & 4.5                                \\ \midrule
\multirow{2}{*}{Llama3-70b} & Others & 1000                               \\
                            & PIE (Ours)             & 4.0 \\ \bottomrule                               
\end{tabular}
\vspace{-0.1in}
\caption{Comparison of average LLM call number for each task.}

\label{callnumber}
\vspace{-0.15in}
\end{table}

\subsubsection{Efficiency and Costs}
Previous works do not consider the cost of calling LLMs. Despite promising performance, they cannot process large-scale graphs. In this section, we will demonstrate high-efficiency and cost-effectiveness of our method from two perspectives:
%\begin{enumerate}[(1)]
%    \item 
    \textbf{(1) Lower single LLM call cost.} Figure~\ref{runtime} presents the average time cost of a single call to LLMs for different algorithms. Clearly, our method exhibits the lowest time consumption across most tasks: with llama3-8b and llama3-70b as LLM backbones, our method reduces the average time cost by 72.8\% and 54.9\% compared to other baselines. This aligns with expectations: other works incorporate serialized graph structures into the prompt, which results in a prompt length of $\mathcal{O}(|E|)$. When processing large-scale graphs, the LLM inference cost is unbearable. In contrast, our approach merely adds pseudocode into prompt, which is independent of the particular graph structure (i.e., prompt length is $\mathcal{O}(1)$). Moreover, the generated executable code can naturally address large-scale graph problems.
    
    %\item 
   \noindent \textbf{(2) Minimal total number of LLM calls.} Furthermore, we investigate the number of times different algorithms call the LLM, as shown in Table~\ref{callnumber}. Other methods need to call the LLM for each test sample, which means the total number equals the test set size (500 small + 500 large). This evidently restricts the application in scenarios involving large-volume data and extensive queries. However, our method requires to call the LLM at most $K$ times for each task. Once the correct executable code is generated, the subsequent process of parsing graph structure and computing answers will be handled by the interpreter. This also implies that our method can be naturally extended to solve large-scale problems.

\begin{table}[t]
\small
\centering
\setlength{\tabcolsep}{0.5mm}
\begin{tabular}{ccccccccccc}
\toprule
\multirow{2}{*}{\begin{tabular}[c]{@{}c@{}}Graph\\ Size \end{tabular}}             & \multirow{2}{*}{\begin{tabular}[c]{@{}c@{}}Method \end{tabular}}        & \multirow{2}{*}{\begin{tabular}[c]{@{}c@{}}CN \end{tabular}}    & \multirow{2}{*}{\begin{tabular}[c]{@{}c@{}}CC \end{tabular}}    & \multirow{2}{*}{\begin{tabular}[c]{@{}c@{}}SD \end{tabular}}    & \multirow{2}{*}{\begin{tabular}[c]{@{}c@{}}GD \end{tabular}}    & \multirow{2}{*}{\begin{tabular}[c]{@{}c@{}}MIS \end{tabular}}  & \multirow{2}{*}{\begin{tabular}[c]{@{}c@{}}MVC \end{tabular}}  & \multirow{2}{*}{\begin{tabular}[c]{@{}c@{}}MCP \end{tabular}}   & \multirow{2}{*}{\begin{tabular}[c]{@{}c@{}}MCS \end{tabular}}  & \multirow{2}{*}{\begin{tabular}[c]{@{}c@{}}TSP  \end{tabular}} \\ \\ \midrule
\multirow{4}{*}{Small} 
                       & w/o pseudo        & 100 & 100 & 100 & 100 & 31 & 30 & 100 & 71 & 0  \\
                       & pseudo-NL        & 100 & 100 & 100 & 100 & 98 & 98 & 100 & 85 & 64 \\
                       & pseudo-Core & 100 & 100 & 100 & 100 & 98 & 56 & 100 & 54 & 42 \\ 
                       \specialrule{0em}{1.0pt}{1.0pt} \cline{2-11}
                               \specialrule{0em}{1.0pt}{1.0pt}
                       & w/ pseudo           & \textbf{100} & \textbf{100} & \textbf{100} & \textbf{100} & \textbf{98} & \textbf{98} & \textbf{100} & \textbf{85} & \textbf{80} \\
                       \midrule
\multirow{4}{*}{Large} 
                       & w/o pseudo        & 100 & 100 & 100 & 100 & 2  & 7  & ---     & ---    & ---    \\
                       & pseudo-NL        & 100 & 100 & 100 & 100 & 87 & 95 & 100 & ---    & 11\\
                       & pseudo-Core & 100 & 100 & 100 & 100 & 87 & 12 & ---     & ---    & 5 \\ 
                       \specialrule{0em}{1.0pt}{1.0pt} \cline{2-11}
                               \specialrule{0em}{1.0pt}{1.0pt}
                       & w/ pseudo           & \textbf{100} & \textbf{100} & \textbf{100} & \textbf{100} & \textbf{87} & \textbf{95} & \textbf{100} & \textbf{28} & \textbf{27} \\ \bottomrule
\end{tabular}
\vspace{-0.1in}
\caption{Accuracy(\%) comparison of different pseudocode prompts. --- means the result of one sample cannot be obtained in 50 seconds. }
\vspace{-0.15in}
\label{diffpsecode}
\end{table}

\subsection{Detailed Analysis}
% In this section, we further analyze the impact of key components of our framework on final performance, including pseudocode prompt and LLM backbone.

% \usepackage{multirow}

\subsubsection{Different Forms of Pseudocode Prompts} Pseudocode prompt plays a crucial role in our method. It provides clear guidance for the generation objectives of the LLM, and directs it to perform code-writing tasks instead of graph understanding. Meanwhile, it prevents LLMs from generating simple brute-force algorithms that cannot be applied to large-scale data. We further investigate the impact of different forms of pseudocode prompts on accuracy, as shown in Table~\ref{diffpsecode}. ``w/o pseudo'' means removing pseudocode prompts. ``pseudo-NL'' denotes that we use LLMs to rephrase pseudocode in natural language, and ``pseudo-Core'' only retains core ideas and removes specific implementation details from ``pseudo-NL''. Examples can be found in Appendix~\ref{different_pseudocode_prompt}. Results show that for polynomial-time tasks, even without pseudocode prompts, LLMs can provide completely correct code. This is because these problems are quite common, and LLMs have been sufficiently trained on them. However, on NP-complete problems, the lack of pseudocode prompts makes LLMs more inclined to output brute-force algorithms that cannot output results within a limited time. Rephrasing pseudocode through LLMs can enhance its readability and improve the robustness of the generated code. However, natural language descriptions are less precise than pseudocode, and missing details lead to wrong results. The significant decrease in the accuracy of ``pseudo-NL'' on MCS and TSP confirms this. Meanwhile, the results of ``pseudo-Core'' also indicate the specific steps in pseudocode prompt are crucial. Otherwise, LLM's output is highly likely to degenerate into a straightforward brute-force algorithm.

\subsubsection{Different LLM Backbones}
For other methods, LLMs with more parameters perform better. However, this distinction is not markedly pronounced in our framework. As delineated in Table~\ref{acc} and Figure~\ref{ar}, the accuracy and approximation rates for llama3-8b and llama3-70b are quite close. Notice that larger-scale LLMs incur more inference cost, which is confirmed by the results in Figure~\ref{runtime}. The only advantage might be that they require fewer trial-and-error attempts to generate correct code, as evidenced in Table~\ref{callnumber}. However, choosing a small-scale LLM as a backbone is sufficient to fulfill our framework.

\subsection{Case Study}
Appendix~\ref{prompt_example_by_ours} and \ref{code_generated_by_pie} present prompts we design for different tasks and executable Python code generated by LLMs. We select TSP for further analysis. Figure~\ref{tspnopse} and Figure~\ref{tsppse} show codes generated by llama3-8b without and with pseudocode. Through careful reading, we observe that without the guide of pseudocode, LLMs tend to output brute-force algorithm (for TSP, it traverses all possible node permutations). Although theoretically ensuring accuracy, it cannot provide answers for large-scale graphs due to extremely long execution time. However, when pseudocode is provided, LLMs compare the complexity of brute-force algorithm with pseudocode and ultimately chooses latter and converts it into executable code, which balances the efficiency and accuracy.

\begin{figure}[t]
\begin{mypython1}
def travelling_salesman(G):
    n = len(G)
    def tsp(i, visited, path, memo):
        if len(visited) == n:
            return path[-1] + G[path[-1]][path[0]]
        if (i, tuple(visited), tuple(path)) in memo:
            return memo[(i, tuple(visited), tuple(path))]
        min_val = float('inf')
        for j in range(n):
            if j not in visited:
                new_visited = visited | {j}
                new_path = path + [j]
                val = path[-1] + G[path[-1]][j] + tsp(j, new_visited, new_path, memo)
                min_val = min(min_val, val)
        memo[(i, tuple(visited), tuple(path))] = min_val
        return min_val
    return tsp(0, {0}, [0], {})  
\end{mypython1}
\vspace{-0.2in}
\caption{Code generated by LLMs without pseudocode.}
\label{tspnopse}
\vspace{-0.15in}
\end{figure}

\begin{figure}[t]
{
\begin{mypython1}
def travelling_salesman(G):
    n = len(G)
    ans = float('inf')
    for start in range(n):
        length = 0
        State = [0] * n
        v = start
        State[v] = 1
        for _ in range(n - 1):
            u = 0
            for i in range(n):
                if State[i] == 0 and (u==0 or G[v][i]<G[v][u]):
                    u = i
            length += G[v][u]
            State[u] = 1
            v = u
        length += G[v][start]
        ans = min(ans, length)
    return ans
\end{mypython1}
}
\vspace{-0.2in}
\caption{Code generated by LLMs with pseudocode.}
\label{tsppse}
\vspace{-0.2in}
\end{figure}

\section{Related Work}
%LLMs have been widely applied across various domains. 
%In the field of graph management, researchers attempt to use LLMs to solve graph computational tasks, and their methods can be classified into the following two groups:
The existing methods using LLMs to solve graph computational tasks can be classified into the following two groups:

\textbf{(1) LLM-based Graph Reasoning using Text-Described Graph Data.}
These approaches directly transform the task and graph structures into text sequences, and then input into LLMs for further reasoning. Many researchers adopt this paradigm to build evaluation datasets for assessing LLMs' reasoning ability on graph tasks. Notable examples include GraphArena~\cite{grapharena}, talk-like-a-graph~\cite{talklikeagraph}, NLGraph~\cite{nlgraph}, GraphInstruct~\cite{graphinstruct} and PSEUDO~\cite{pse}. GraphArena extracts test graphs from real-world data and assigns real-world meanings to each node before transforming them into text sequences. Talk-like-a-graph  defines graph encoder function in natural language, while NLGraph incorporates build-a-graph prompt~\cite{cot} and the core algorithmic idea into the input to enhance the reasoning ability of LLMs. Building upon previous explorations, GraphInstruct proposes three methods to describe graph structures and two approaches to handle node IDs. PSEUDO goes a step further by attempting to incorporate pseudocode as prior knowledge into the prompt and guides LLMs to perform reasoning in accordance with algorithmic cues. In addition to these training-free methods, researchers also try to fine-tune LLMs to expect better results, such as GraphWiz~\cite{graphwiz} uses mixed-task instruction tuning and DPO~\cite{dpo} alignment to teach LLMs to develop a logical reasoning process that follows human problem-solving behavior. 

% Notable examples include GraphArena~\cite{grapharena}, NLGraph~\cite{nlgraph}, PSEUDO~\cite{pse} and talk-like-a-graph~\cite{talklikeagraph}, which have laid the foundation for evaluating LLMs' graph task reasoning performance. However, the experimental results also highlight the need for improving LLMs' understanding of graph data during the reasoning process, as well as the significant impact of graph data scale on LLMs' graph reasoning performance. To address these shortcomings, some researchers have employed training and fine-tuning methods to inject knowledge into LLMs, enabling them to better complete reasoning tasks by understanding graph data and graph tasks. For instance, GraphWiz uses a two-stage LLM fine-tuning alignment paradigm of LoRA and DPO to train the model on nine graph tasks, thereby enhancing its understanding of graph data and specific graph tasks. Similarly, GraphInstruct{~\cite{graphinstruct}} employs LoRA fine-tuning and label masking training strategies to improve LLMs' graph reasoning performance. However, the enhanced models from these research efforts have relatively weak generalization capabilities. For example, although GraphWiz{~\cite{graphwiz}} performs well on the training and testing datasets constructed by the research team, its accuracy on new benchmarks like GraphArena is only 0.2\% to 1.2\%. Moreover, due to the context window limitations of LLMs and the hallucination problem, LLM graph reasoning is inevitably affected when the scale of graph data is large.

\textbf{(2) LLM-based Graph Reasoning via Code Generation.}
%For these methods, the prompt describes the graph task, and LLMs are responsible to generate codes to solve the corresponding graph task. 
During the reasoning process, LLMs act as a solution provider, which means they no longer rely on specific graph structure but propose a unified solution for the same category of graph tasks. For instance, GraphEval2000~\cite{grapheval} leverages 40 graph-related tasks from LeetCode to assess LLMs' problem-solving ability based on programming. GraphTeam~\cite{graphteam} integrates multiple LLM-based agents to collaboratively write code for graph reasoning tasks, including search, question, coding, reasoning, and answer agents. GCoder~\cite{gcoder} employs SFT and Reinforcement Learning from Compiler Feedback to fine-tune LLMs for generating preferred code. However, these methods are still constrained by the high costs of repeatedly calling LLMs within multi-agent systems and finetuning LLMs.

% \subsection{Prompt Engineering Enhances LLMs Inference}

% The versatility of LLMs enables their extensive application across various domains, such as question answering, programming, writing, and abstract reasoning, by leveraging knowledge from large-scale and multi-domain corpora. To enhance the performance of LLMs inference within a specific domain, numerous strategies have been proposed, such as Retrieval-Augmented Generation (RAG)~\cite{gao2024retrievalaugmentedgenerationlargelanguage}, domain fine-tuning, and prompt engineering. Notably, well-designed prompts have been shown to significantly improve reasoning capabilities in LLMs, with techniques like Chain-of-Thought (CoT)~\cite{cot} prompting and few-shot In-context learning being validated across several fields.

% {In the context of graph reasoning for LLMs, researchers have employed CoT and few-shot prompt designs, such as GraphArena, to mitigate the hallucination problem often encountered in LLMs reasoning. Furthermore, GraphEval2000 introduces a novel prompt design framework, Structured Symbol Decomposition (SSD), aimed at enhancing the programming performance of LLMs when addressing complex graph-related problems. This paper presents a comparative analysis of SSD and CoT, assessing the effectiveness of the SSD approach. Additionally, PSEUDO proposes pseudocode hints tailored for graph problems to assist LLMs in more effectively resolving inference challenges related to graph data.

\section{Conclusion}
In this paper, we highlight two limitations of existing methods that use LLMs to solve graph computational tasks: low accuracy caused by LLMs' misinterpretation of graph structure and prohibitive inference cost of LLMs due to excessively long prompts. We propose a new framework PIE, which instructs LLMs to understand the task description and write code. To assist LLMs in generation, we introduce pseudocode injection method and employ trial-and-error to select the optimal code. Experiments show that PIE outperforms other methods in accuracy and has the lowest inference cost.

\clearpage

% \clearpage
\bibliographystyle{named}
\bibliography{ijcai25}

\clearpage
\appendix

\onecolumn
\begin{center}
\textbf{\LARGE Appendix}
% \vspace{2em}
\end{center}

\section{Experimental Results of DeepSeek-V3 as LLM Backbone}
\label{deepseek_exp}

\begin{figure}[ht]
\vspace{-0.2in}
    \centering
    \subfloat[DeepSeek-V3 small graph]{\includegraphics[width=0.22\linewidth]{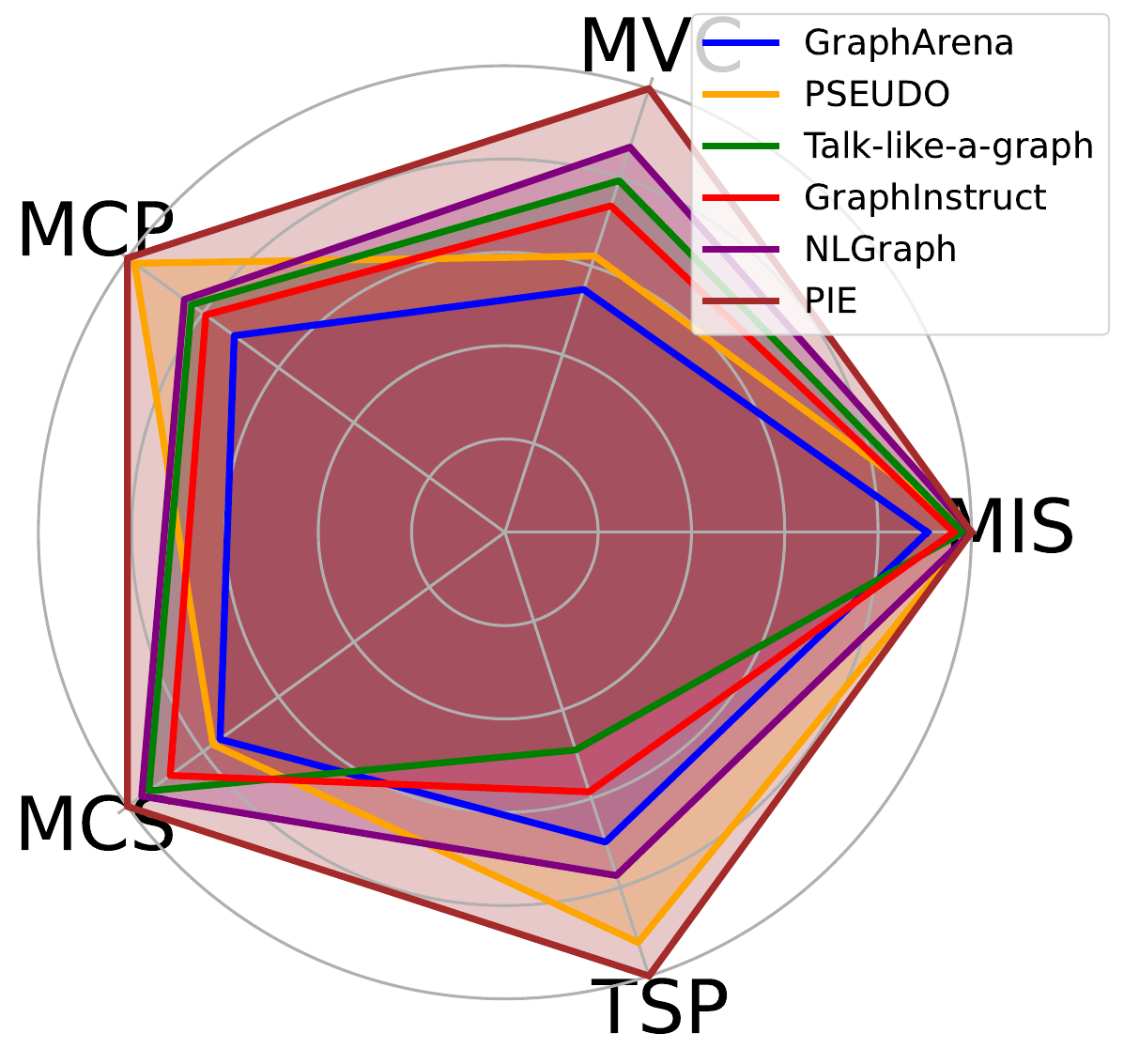}}%
    \label{feasible-deepseek-small}
    \subfloat[DeepSeek-V3 large graph]{\includegraphics[width=0.22\linewidth]{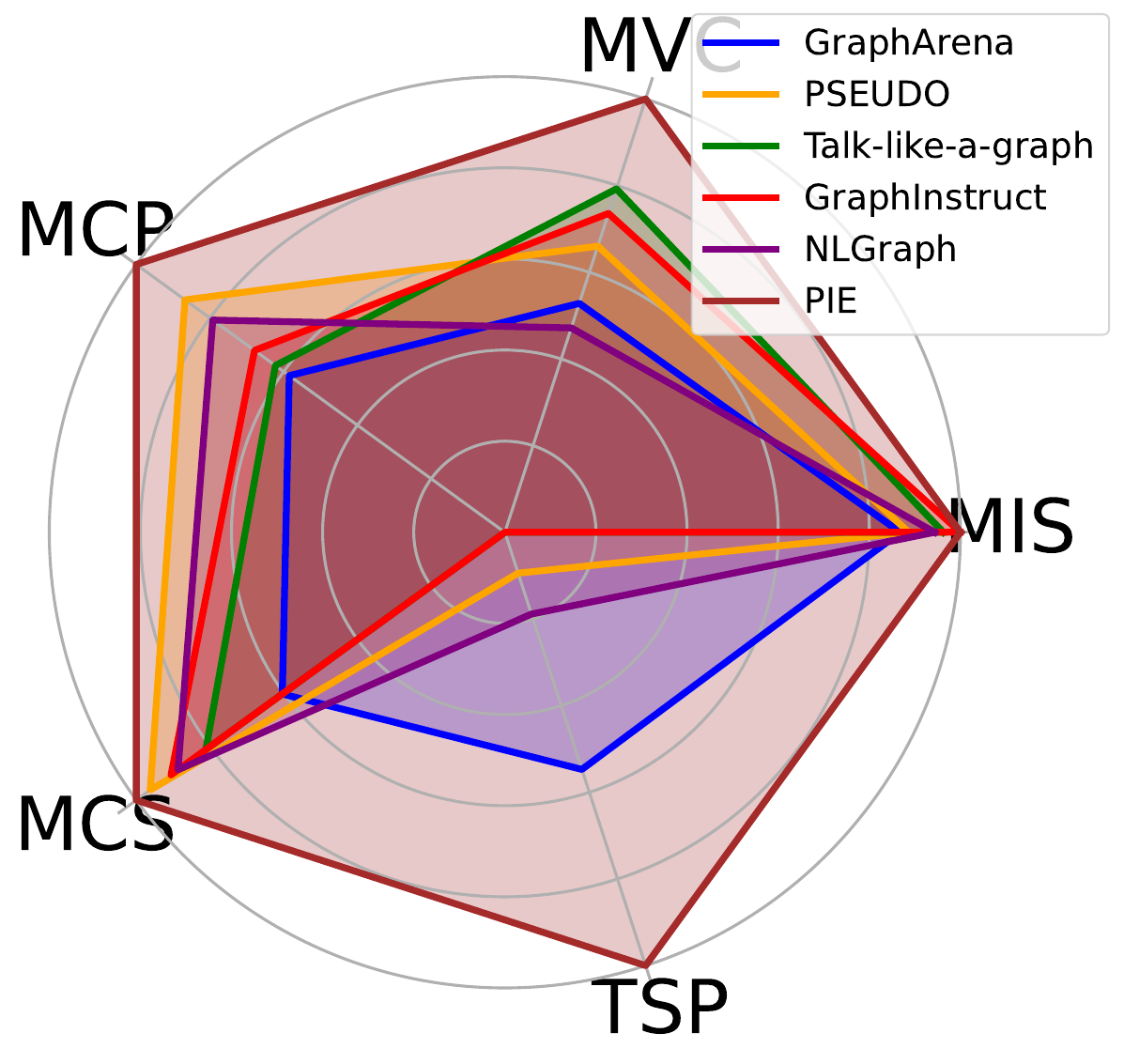}}%
    \label{feasible-deepseek-large}
    \vspace{-0.1in}
    \caption{Feasible rate comparison of various methods on different NP-complete tasks with LLM backbone DeepSeek-V3. Higher is better.}
    \label{deepseek-fr}
    \vspace{-0.15in}
\end{figure}

\begin{figure}[ht]
\vspace{-0.15in}
    \centering
    \subfloat[DeepSeek-V3 small graph]{\includegraphics[width=0.49\linewidth]{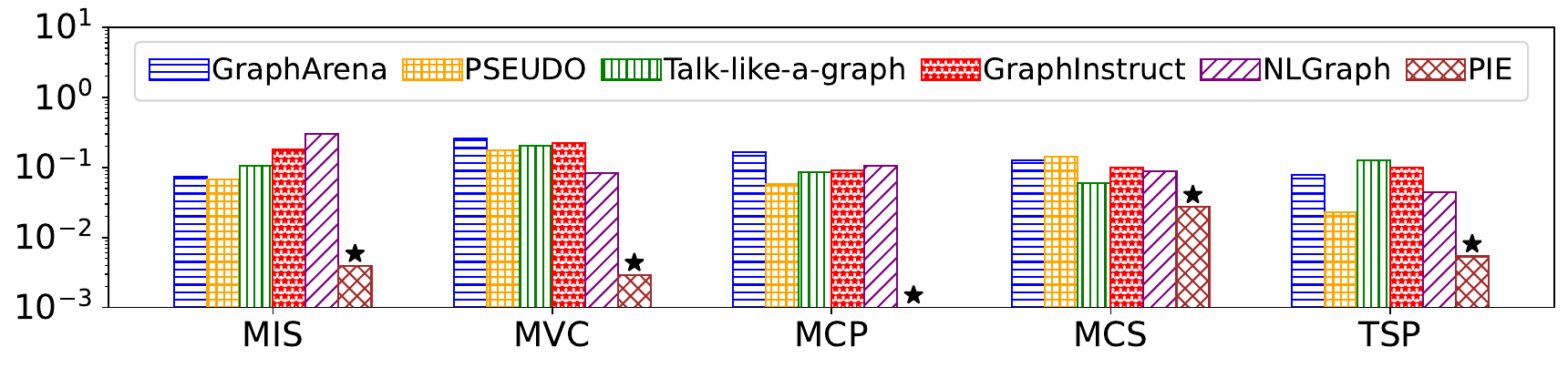}}%
    \label{ratio-deepseek-small}
    % \vspace{-0.1in}
    \subfloat[DeepSeek-V3 large graph]{\includegraphics[width=0.49\linewidth]{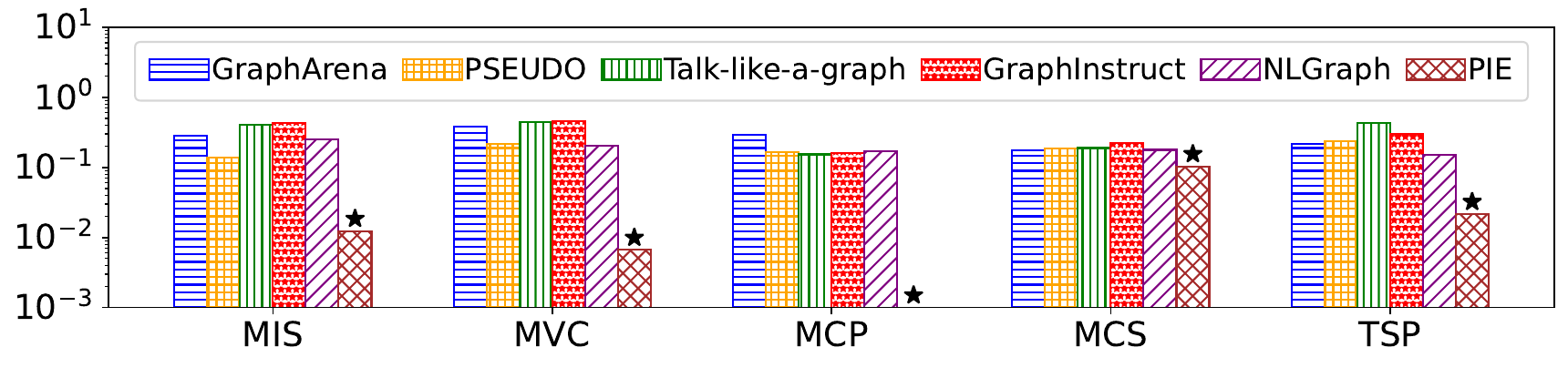}}%
    \label{ratio-deepseek-large}
    \vspace{-0.1in}
    \caption{Approximation ratio comparison of various methods on different NP-complete tasks with LLM backbone DeepSeek-V3. Lower is better.}
    \label{deepseek-ar}
    \vspace{-0.2in}
\end{figure}

\begin{figure*}[t]
\vspace{-0.1in}
    \centering
    {\includegraphics[width=0.85\linewidth]{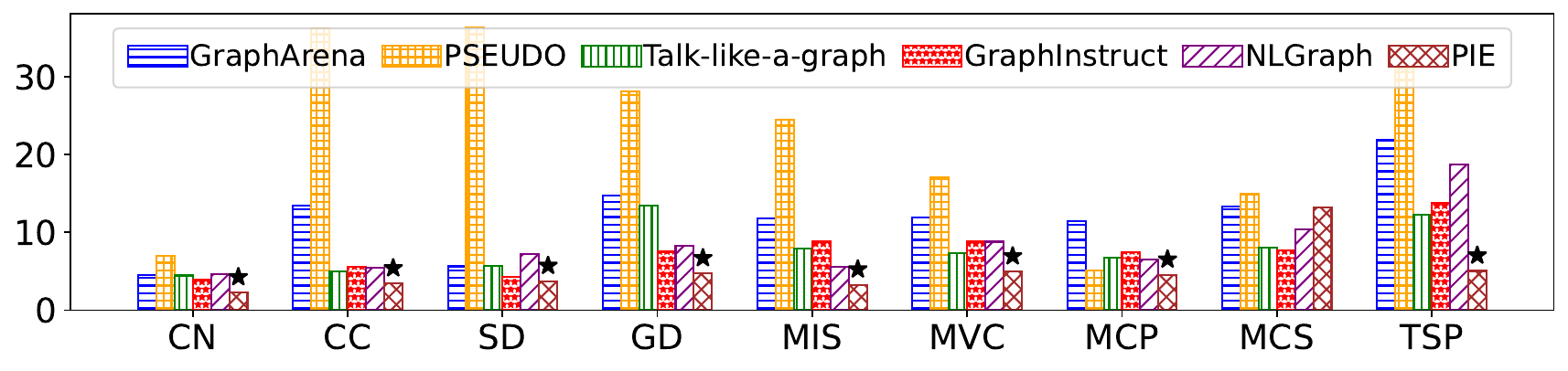}}%
    \label{deepseek-time}
    \vspace{-0.1in}
    \caption{Time cost comparison for a single call to DeepSeek-V3.}
    \label{deepseek-runtime}
    \vspace{-0.1in}
\end{figure*}

In this section, we further analyze the impact of using the latest DeepSeek-V3 as the LLM backbone on the performance of baselines and our method. From Table~\ref{deepseekacc}, we observe that the accuracy of other methods has improved compared to llama3-8b and llama3-70b. However, they still fall short of our approach. From Figure~\ref{deepseek-fr} and \ref{deepseek-ar}, our method PIE still ensures 100\% feasibility of output solutions and a low approximation ratio, which guarantees the practicality of our approach. From an efficiency perspective, Figure~\ref{deepseek-runtime} demonstrates that PIE has the advantage of the lowest single-call LLM cost. At the same time, our method eliminates the need for repeatedly calling LLMs, since the generated code can be reused for tasks of the same category. In contrast, other methods require calling the LLM for new test samples, resulting in significant inference costs. This is also corroborated by Table ~\ref{deepseekcallnumber}.

\begin{table}[ht]
\small
\centering
\setlength{\tabcolsep}{1mm}
\begin{tabular}{ccccccccccc}
\toprule

                            &                          & \multicolumn{4}{c}{Polynomial-time Tasks}                                                                                       & \multicolumn{5}{c}{NP-complete Tasks}        \\ \specialrule{0em}{1.0pt}{1.0pt} \cline{3-11}
                               \specialrule{0em}{1.0pt}{1.0pt}
\multirow{-2}{*}{LLM}       & \multirow{-2}{*}{Method} & CN                           & CC                           & SD                           & GD                           & MIS   & MVC   & MCP    & MCS   & TSP   \\
\midrule
                            & GraphArena               & \small{79.2/56.6} & \small{96.2/45.2} & \small{81.1/43.4} & \small{73.5/24.5} & \small{69.8/3.7}  & \small{43.4/20.7}  & \small{62.6/24.5}   & \small{50.9/16.9}  & \small{16.9/0.0}  \\
                            & PSEUDO                   
                            & \small{90.6/84.9} & \small{98.1/60.4} & \small{100.0/56.6} & \small{73.6/32.1}                        & \small{88.7/56.6} & \small{56.6/35.9} & \small{73.6/32.1}  & \small{47.2/5.7} & \small{58.5/0.0}  \\
                            & Talk-like-a-graph        & \small{96.2/100.0} & \small{94.3/71.7} & \small{92.4/60.4}                        & \small{77.4/35.9}                        & \small{69.8/3.8} & \small{52.8/7.6} & \small{71.7/28.3}  & \small{77.3/7.6} & \small{45.2/0.0}  \\
                            & GraphInstruct            & \small{92.4/88.7} & \small{90.6/39.6}                        & \small{100.0/64.2}                        & \small{54.7/45.3}                        & \small{43.4/3.8} & \small{39.6/11.3} & \small{73.6/30.2}  & \small{60.4/9.4} & \small{52.8/0.0}  \\
                            & NLGraph                  & \small{88.7/75.4}                        & \small{88.7/54.7}                        & \small{90.6/45.3}                        & \small{73.6/37.7}                        & \small{30.2/15.1} & \small{71.7/32.1} & \small{60.4/24.5}  & \small{62.3/24.5} & \small{52.8/0.0}  \\ \specialrule{0em}{1.0pt}{1.0pt} \cline{2-11}
                               \specialrule{0em}{1.0pt}{1.0pt}
\multirow{-6}{*}{\begin{tabular}[c]{@{}l@{}}DeepSeek\\ \quad V3\end{tabular}} & PIE (Ours)                     & \small{\textbf{100.0/100.0}}                       & \small{\textbf{100.0/100.0}}                       & \small{\textbf{100.0/100.0}}                       & \small{\textbf{100.0/100.0}}                       & \small{\textbf{98.0/87.6}} & \small{\textbf{98.8/95.2}} & \small{\textbf{100.0/100.0}} & \small{\textbf{88.0/29.3}} & \small{\textbf{80.2/27.4}} \\ \bottomrule

\end{tabular}
\vspace{-0.1in}
\caption{Accuracy(\%) comparison among different methods on small/large graphs with LLM backbone DeepSeek-V3. Higher is better.}
\vspace{-0.1in}
\label{deepseekacc}
\end{table}

\begin{table}[t]
\small
\centering
\setlength{\tabcolsep}{5mm}
\begin{tabular}{cccccccccc}
\toprule
Method    & CN   & CC   & SD   & GD   & MIS  & MVC  & MCP  & MCS  & TSP  \\ \midrule
Others    & 1000 & 1000 & 1000 & 1000 & 1000 & 1000 & 1000 & 1000 & 1000 \\
PIE (Ours) & 10   & 1    & 1    & 1    & 10   & 1    & 1    & 10   & 10  \\ \bottomrule
\end{tabular}
\vspace{-0.1in}
\caption{Comparison of the number of LLM calls for different tasks.}
\vspace{-0.1in}
\label{deepseekcallnumber}
\end{table}

\section{Prompt Examples Used by Other Methods}
\label{prompt_example_by_others}
We provide prompt examples for other baselines in our experiments. We use MVC problem as an example.

\subsection{System prompt}

\begin{tcolorbox}[colback=gray!10, colframe=black, rounded corners, boxrule=1.5pt, fontupper=\normalsize, left=2mm, right=2mm, top=1mm, bottom=1mm]

You are an expert in the field of graph algorithms and are currently solving the minimum vertex cover problem. I will provide a series of problems, and please try to solve these problems step by step and give the final answers in the following format: ``The answer is [number]''. Please ensure that the output in the last line is ``The answer is [number]''. [number] is filled with your answer.

\end{tcolorbox}

\subsection{GraphArena}

\begin{tcolorbox}[colback=gray!10, colframe=black, rounded corners, boxrule=1.5pt, fontupper=\normalsize, left=2mm, right=2mm, top=1mm, bottom=1mm]

Your task is to solve the Minimum Vertex Cover problem in the given social network. In this network, each node represents a user, and each edge represents a friendship connection. You need to identify the smallest subset of users such that every friendship connection has at least one user from this subset.
\newline \newline **Problem to Solve** \newline \newline - Users in the network: Pamela Haynes, Kyle Meadows, Adam Nichols, Anna Lowery, Heather Dixon, Matthew Lee, Elizabeth Wood, Stephen Hess.\newline- Fiendship connections: Pamela Haynes and Stephen Hess, Kyle Meadows and Matthew Lee, Kyle Meadows and Stephen Hess, Kyle Meadows and Adam Nichols, Adam Nichols and Stephen Hess, Adam Nichols and Heather Dixon, Anna Lowery and Stephen Hess, Heather Dixon and Stephen Hess, Matthew Lee and Stephen Hess, Elizabeth Wood, and Stephen Hess.\newline Present the size of this undirected graph's minimum vertex cover. Please just output the number.
\end{tcolorbox}

\subsection{PSEUDO}

\begin{tcolorbox}[colback=gray!10, colframe=black, rounded corners, boxrule=1.5pt,fontupper=\normalsize, left=2mm, right=2mm, top=1mm, bottom=1mm]

Pseudocode is as follows:
\begin{verbatim}
Function find_Minimum_Vertex_Cover(G):
    vertex_cover ← empty set
    edges ← empty set
    for each vertex and its neighbors in G:
        for each neighbor:
            if (vertex, neighbor) is not in edges and
                (neighbor, vertex) is not in edges:
                add (vertex, neighbor) to edges
    while edges is not empty:
        max_coverage = -1
        best_vertex = None
        for each vertex and its neighbors in G:
            covered_edges = {edge | edge in edges and vertex in edge}
            coverage = size of covered_edges
            if coverage > max_coverage:
                max_coverage = coverage
                best_vertex = vertex
        add best_vertex to vertex_cover
        edges = edges - {edge | edge in edges and best_vertex in edge}
    return len(vertex_cover)
\end{verbatim}
Let G be an undirected graph. G has nodes 0,1,2,3,4,5,6,7. The edgelist of graph G is the following: [(0, 5), (0, 1), (1, 7), (1, 5), (1, 2), (1, 6), (1, 3), (1, 4), (2, 5), (2, 3)]. Each edge appears only once. What is the size of the minimum vertex cover in this undirected graph? Please only output the number. Follow the provided pseudocode step-by-step.
\end{tcolorbox}

\subsection{Talk-like-a-graph}

\begin{tcolorbox}[colback=gray!10, colframe=black, rounded corners, boxrule=1.5pt, fontupper=\normalsize, left=2mm, right=2mm, top=1mm, bottom=1mm]

G0 describes a graph among nodes 0,1,2,3,4,5,6,7.\newline
In this graph:\newline
Node 0 is connected to nodes 5,1.\newline
Node 1 is connected to nodes 0,7,5,2,6,3,4.\newline
Node 2 is connected to nodes 1,5,3.\newline
Node 3 is connected to nodes 1,2.\newline
Node 4 is connected to nodes 1.\newline
Node 5 is connected to nodes 0,1,2.\newline
Node 6 is connected to nodes 1.\newline
Node 7 is connected to nodes 1.\newline
Question: What is the size of the minimum vertex cover in this undirected graph? Please only output the number.

\end{tcolorbox}

\subsection{GraphInstruct}

\begin{tcolorbox}[colback=gray!10, colframe=black, rounded corners, boxrule=1.5pt, fontupper=\normalsize, left=2mm, right=2mm, top=1mm, bottom=1mm]

Given an undirected graph:\newline
Node 0 is connected to nodes 5,1.\newline
Node 1 is connected to nodes 7,5,2,6,3,0,4.\newline
Node 2 is connected to nodes 5,1,3.\newline
Node 3 is connected to nodes 2,1.\newline
Node 4 is connected to nodes 1.\newline
Node 5 is connected to nodes 0,1,2.\newline
Node 6 is connected to nodes 1.\newline
Node 7 is connected to nodes 1.\newline
What is the size of the minimum vertex cover in this undirected graph? Please only output the number.

\end{tcolorbox}

\subsection{NLGraph}

\begin{tcolorbox}[colback=gray!10, colframe=black, rounded corners, boxrule=1.5pt, fontupper=\normalsize, left=2mm, right=2mm, top=1mm, bottom=1mm]

The core idea of the algorithm can be summarized as follows:\newline
**Initialize**: Start with an empty vertex cover and an empty set of edges.\newline
**Collect Edges**: Iterate through the graph to populate the set of unique edges.\newline
**Greedy Selection**: Repeatedly select the vertex that covers the most uncovered edges.\newline
Update: Add the selected vertex to the vertex cover and remove all edges it covers.\newline
**Terminate**: Continue until all edges are covered, then return the size of the vertex cover.\newline
This algorithm employs a greedy strategy to approximate the minimum vertex cover by prioritizing vertices that cover the maximum number of edges at each step.\newline
\newline
In an undirected graph, the nodes are numbered from 0 to 7. The edges are:an edge between node 0 and node 5, an edge between node 0 and node 1, an edge between node 1 and node 7, an edge between node 1 and node 5, an edge between node 1 and node 2, an edge between node 1 and node 6, an edge between node 1 and node 3, an edge between node 1 and node 4, an edge between node 2 and node 5, an edge between node 2 and node 3.\newline Let's construct a graph with the nodes and edges first. \newline Question: What is the size of the minimum vertex cover in this undirected graph? Please only output the number.
\end{tcolorbox}

\section{Prompt Examples Used by Our Approach PIE}
\label{prompt_example_by_ours}
In this section, we provide the designed prompts for PIE, which consists of three parts: system prompt, problem prompt, and pseudocode prompt.

\subsection{System prompt}

\begin{tcolorbox}[colback=gray!10, colframe=black, rounded corners, boxrule=1.5pt, fontupper=\normalsize, left=2mm, right=2mm, top=1mm, bottom=1mm]

As an expert in the field of graph algorithms, you will take on the role of interpreting descriptions of graph algorithms provided by the user. Based on these descriptions, you will abstract the problem into a graph algorithm issue and provide a Python code snippet that can be executed correctly. Please ensure that the Python code is executable without any comments. Please only output Python functions starting with def, without any other information, including test cases and comments.

\end{tcolorbox}

\subsection{Problem prompt}
\subsubsection{Common Neighbors}
\begin{tcolorbox}[colback=gray!10, colframe=black, rounded corners, boxrule=1.5pt, fontupper=\normalsize, left=2mm, right=2mm, top=1mm, bottom=1mm]
Given a directed graph G with no weights on the edges, calculate the common neighbors of node pairs in the graph. The Python code will receive the graph G and two nodes u, v. The graph G is passed in the format of an adjacency list: G[i] stores the neighbors of i. All elements are integers, and the graph G is numbered starting from 0.\newline\newline
        The algorithm process can be described as follows: \{PSEUDOCODE INSERT HERE\} \newline\newline
        Please complete the following Python code according to the provided algorithm description so that the function returns a list representing the common neighbors of nodes u and v in graph G. def common\_neighbors(G, u, v):

\end{tcolorbox}

\subsubsection{Connected Components}
\begin{tcolorbox}[colback=gray!10, colframe=black, rounded corners, boxrule=1.5pt, fontupper=\normalsize, left=2mm, right=2mm, top=1mm, bottom=1mm]

Given an undirected graph G with no weights on the edges, calculate the number of connected components in the graph. A connected component X in an undirected graph is defined as a set of nodes such that any two nodes in X are connected by a path within X. The Python code will receive the graph G, which is passed in the format of an adjacency list: G[i] stores the neighbors of node i. All elements are integers, and the graph G is numbered starting from 0.\newline\newline
        The algorithm process can be described as follows: \{PSEUDOCODE INSERT HERE\} \newline\newline
        Please complete the following Python code according to the provided algorithm description so that the function returns a value representing the number of connected components in the undirected graph G. def connected\_component\_undirected(G):

\end{tcolorbox}

\subsubsection{Shortest Path}
\begin{tcolorbox}[colback=gray!10, colframe=black, rounded corners, boxrule=1.5pt, fontupper=\normalsize, left=2mm, right=2mm, top=1mm, bottom=1mm]

"Given a directed graph G with no weights on the edges, and two nodes u and v, calculate the length of the shortest path between the two points. The Python code will receive the graph G and nodes u, v. The graph G is passed in the format of an adjacency list. All elements are integers, and the graph G is numbered starting from 0.\newline\newline
        The algorithm process can be described as follows: \{PSEUDOCODE INSERT HERE\} \newline\newline
        Please complete the following Python code according to the provided algorithm description so that the function returns a value representing the length of the shortest path between u and v. def shortest\_path\_unweighted(G, u, v):
        
\end{tcolorbox}

\subsubsection{Graph Diameter}
\begin{tcolorbox}[colback=gray!10, colframe=black, rounded corners, boxrule=1.5pt, fontupper=\normalsize, left=2mm, right=2mm, top=1mm, bottom=1mm]

Given an undirected graph G with no weights on the edges, calculate the diameter of the graph. The diameter of a graph is defined as the maximum length of the shortest path between any two vertices in the graph. The Python code will receive the graph G, where the graph G is passed in the format of an adjacency list: G[i] stores the neighbors of node i. All elements are integers, and the graph G is numbered starting from 0.\newline\newline
        The algorithm process can be described as follows: \{PSEUDOCODE INSERT HERE\} \newline\newline
        Please complete the following Python code according to the provided algorithm description so that the function returns a value representing the diameter of graph G. def graph\_diameter(G):
        
\end{tcolorbox}

\subsubsection{Maximum Independent Set}
\begin{tcolorbox}[colback=gray!10, colframe=black, rounded corners, boxrule=1.5pt, fontupper=\normalsize, left=2mm, right=2mm, top=1mm, bottom=1mm]

Given an undirected graph G with no weights on the edges, calculate the size of the maximum independent set on the graph. An independent set X on the graph is defined as a set of vertices such that no two vertices u, v in X are adjacent. The Python code will receive the graph G, where the graph G is passed in the format of an adjacency list: G[i] stores the neighbors of i. All elements are integers, and the graph G is numbered starting from 0.\newline\newline
        The algorithm process can be described as follows: \{PSEUDOCODE INSERT HERE\} \newline\newline
        Please complete the following Python code according to provided algorithm description so that the function returns a value representing the size of the maximum independent set of the undirected graph G. def maximum\_independent\_set(G):
        
\end{tcolorbox}

\subsubsection{Minimum Vertex Cover}
\begin{tcolorbox}[colback=gray!10, colframe=black, rounded corners, boxrule=1.5pt, fontupper=\normalsize, left=2mm, right=2mm, top=1mm, bottom=1mm]

Given an undirected graph G with no weights on the edges, calculate the minimum vertex cover problem of the graph. A vertex cover X of a graph G is defined as a set of vertices such that every edge of G has at least one endpoint in X. The Python code will receive the graph G, where the graph G is passed in the format of an adjacency list: G[i] stores the neighbors of i. All elements are integers, and the graph G is numbered starting from 0.\newline\newline
        The algorithm process can be described as follows: \{PSEUDOCODE INSERT HERE\} \newline\newline
        Please complete the following Python code according to the provided algorithm description so that the function returns a value representing the size of the minimum vertex cover of the undirected graph G. def minimum\_vertex\_cover(G):
        
\end{tcolorbox}

\subsubsection{Maximum Clique Problem}
\begin{tcolorbox}[colback=gray!10, colframe=black, rounded corners, boxrule=1.5pt, fontupper=\normalsize, left=2mm, right=2mm, top=1mm, bottom=1mm]

Given an undirected graph G with no weights on the edges, calculate the maximum clique problem of the graph. A clique X of a graph G is defined as a set of vertices such that there is an edge between any two nodes u, v in X. The Python code will receive the graph G, where the graph G is passed in the format of an adjacency list: G[i] stores the neighbors of i. All elements are integers, and the graph G is numbered starting from 0.\newline\newline
        The algorithm process can be described as follows: \{PSEUDOCODE INSERT HERE\} \newline\newline
        Please complete the following Python code according to the provided algorithm description so that the function returns a value representing the size of the maximum clique of the undirected graph G. def maximum\_clique(G):
        
\end{tcolorbox}

\subsubsection{Maximum Common Subgraph}
\begin{tcolorbox}[colback=gray!10, colframe=black, rounded corners, boxrule=1.5pt, fontupper=\normalsize, left=2mm, right=2mm, top=1mm, bottom=1mm]

Given two undirected graphs G1, G2 with no weights on the edges and no labels on the nodes, calculate the maximum common induced subgraph of G1 and G2. The Python code will receive two graphs G1 and G2, where both of them are passed in the format of networkx.Graph. All elements are integers, and G1, G2 are numbered starting from 0.\newline\newline
        The algorithm process can be described as follows: \{PSEUDOCODE INSERT HERE\} \newline\newline
        Please complete the following Python code according to the provided algorithm description so that the function returns a value representing the size of the maximum common subgraph of G1 and G2. def maximum\_common\_subgraph(G1, G2):
        
\end{tcolorbox}

\subsubsection{Traveling Salesman Problem}
\begin{tcolorbox}[colback=gray!10, colframe=black, rounded corners, boxrule=1.5pt, fontupper=\normalsize, left=2mm, right=2mm, top=1mm, bottom=1mm]

Given an undirected complete graph G where every pair of nodes is connected by an edge, and each edge has a weight, calculate the Traveling Salesman Problem (TSP) on the graph, which is the shortest cycle that visits each node exactly once and returns to the starting node. The Python code will receive the graph G, where the graph G is passed in the format of an adjacency matrix: G[i][j] stores the weight of edge (i, j). All elements are integers, and the graph G is numbered starting from 0.\newline\newline
        The algorithm process can be described as follows: \{PSEUDOCODE INSERT HERE\} \newline\newline
        Please complete the following Python code so that the function returns a value representing the solution to the Traveling Salesman Problem on the undirected graph G, i.e., the minimum edge weight sum of the cycle. def travelling\_salesman(G):
        
\end{tcolorbox}

\subsection{Pseudocode prompt}
\subsubsection{Common Neighbors}
\begin{tcolorbox}[colback=gray!10, colframe=black, rounded corners, boxrule=1.5pt, fontupper=\normalsize, left=2mm, right=2mm, top=1mm, bottom=1mm]
\begin{verbatim}
Function find_common_neighbors(G, node1, node2):
    create an empty list common_neighbors
    Iterate all node in G:
        if node is in both node1's and node2's neighbors:
            add node to common_neighbors
    return common_neighbors
\end{verbatim}
\end{tcolorbox}

\subsubsection{Connected Components}
\begin{tcolorbox}[colback=gray!10, colframe=black, rounded corners, boxrule=1.5pt, fontupper=\normalsize, left=2mm, right=2mm, top=1mm, bottom=1mm]
\begin{verbatim}
Function find_all_connected_components(G):
    V ← G.nodes, V' ← empty set, K ← 1
    result ← List()
    while V' != V:
        select s from V - V'
        add s to V'
        Q ← {s}, C_K ← {s}
        while Q is not empty:
            u ← Q.dequeue()
            for each v in u's neighbors:
                if v is not in V': 
                    add v to V', Q.enqueue(v), add v to C_K
        add C_K to result
        K ← K + 1
    return K - 1
\end{verbatim}

\end{tcolorbox}

\subsubsection{Shortest Path}
\begin{tcolorbox}[colback=gray!10, colframe=black, rounded corners, boxrule=1.5pt, fontupper=\normalsize, left=2mm, right=2mm, top=1mm, bottom=1mm]
\begin{verbatim}
Function find_shortest_path(G, start, end):
    Initialize queue Q: Q ← [(start, [start], 0)]
    Create set of visited nodes: V' ← {}
    While Q is not empty:
        (node, path, distance) ← Q.dequeue()
        If node is the end node:
            result ← path
            return distance
        Mark node as visited: add node to V'
        Iterate over all neighbors of node
            If neighbor is not in V':
               Enqueue (neighbor, list(path) + [neighbor], distance + 1) to Q
\end{verbatim}
\end{tcolorbox}

\subsubsection{Graph Diameter}
\begin{tcolorbox}[colback=gray!10, colframe=black, rounded corners, boxrule=1.5pt, fontupper=\normalsize, left=2mm, right=2mm, top=1mm, bottom=1mm]
\begin{verbatim}
Function find_graph_diameter(G):
    diameter ← 0
    Iterate over all nodes in graph
    visited ← set(), distance ← {node: float('inf') for node in graph}
    Initialize queue Q with start node: Q ← [start]
    distance[start] = 0
    while Q is not empty do:
        current_node ← Q.dequeue()
        add current_code to visited
        Iterate over all neighbors of current_node
            If neighbor is not in visited:
                add neighbor to visited
                distance[neighbor] ← distance[current_node] + 1
                Enqueue neighbor to Q 
    longest_path ← max(distance.values())
    diameter ← max{diameter, longest_path}
    return diameter
\end{verbatim}
\end{tcolorbox}

\subsubsection{Maximum Independent Set}
\begin{tcolorbox}[colback=gray!10, colframe=black, rounded corners, boxrule=1.5pt, fontupper=\normalsize, left=2mm, right=2mm, top=1mm, bottom=1mm]
\begin{verbatim}
Function find_maximum_independent_set(G):
    All nodes are re-ordered from smallest to largest by degree.
    IS ← empty set
    State[v] ← 0 for each node in G 
    for each node v in V do:
        if State[v] == 0:
            add v to IS
            for each u in v's neighbor do:
                if State[u] == 0:
                    State[u] ← 1
    return length of IS
\end{verbatim}
\end{tcolorbox}

\subsubsection{Minimum Vertex Cover}
\begin{tcolorbox}[colback=gray!10, colframe=black, rounded corners, boxrule=1.5pt, fontupper=\normalsize, left=2mm, right=2mm, top=1mm, bottom=1mm]
\begin{verbatim}
Function find_Minimum_Vertex_Cover(G):
    vertex_cover ← empty set
    edges ← empty set
    for each vertex and its neighbors in G:
        for each neighbor:
            if (vertex, neighbor) is not in edges and
                (neighbor, vertex) is not in edges:
                add (vertex, neighbor) to edges
    while edges is not empty:
        max_coverage = -1
        best_vertex = None
        for each vertex and its neighbors in G:
            covered_edges = {edge | edge in edges and vertex in edge}
            coverage = size of covered_edges
            if coverage > max_coverage:
                max_coverage = coverage
                best_vertex = vertex
        add best_vertex to vertex_cover
        edges = edges - {edge | edge in edges and best_vertex in edge}
    return len(vertex_cover)
\end{verbatim}
\end{tcolorbox}

\subsubsection{Maximum Clique Problem}
\begin{tcolorbox}[colback=gray!10, colframe=black, rounded corners, boxrule=1.5pt, fontupper=\normalsize, left=2mm, right=2mm, top=1mm, bottom=1mm]
\begin{verbatim}
Function find_maximum_clique(G): 
    Initial max_cliques C ← list()
    
    Function bron_kerbosch_recursive(r, p, x):
        if p and x are both empty:
            C ← C.append(r)
        Iterate over all vertex v in p.copy():
            new_r ← r.union(v)
            new_p ← p.intersect(v's neighbor) 
            new_x ← x.intersect(v's neighbor)
            bron_kerbosch_recursive(new_r, new_p, new_x)
            p ← p.remove(v)
            x ← x.add(v)
            
    bron_kerbosch_recursive(set(), set(G.keys()), set())
    return max_clique of C
\end{verbatim}
\end{tcolorbox}

\subsubsection{Maximum Common Subgraph}
\begin{tcolorbox}[colback=gray!10, colframe=black, rounded corners, boxrule=1.5pt, fontupper=\normalsize, left=2mm, right=2mm, top=1mm, bottom=1mm]
\begin{verbatim}
Function find_maximum_common_subgraph(G1,G2):
    # Initialization:
    Create H_nodes to store matched node pair (u1,u2).
    Create matched_nodes1 to store matched nodes in G1.
    Create matched_nodes2 to store matched nodes in G2.
    Create a dictionary node_map to map nodes from G1 to G2.
    Create a dictionary node_map_reverse to map nodes from G2 to G1.

    # Calculate Node Degree Differences and Sort:
    tot_degree_diff ← []
    for each pair (u1,u2) in V1×V2 {
        degree_diff ← |degree(u1) - degree(u2)|
        add (degree_diff, u1, u2) to tot_degree_diff
    }
    sort tot_degree_diff in ascending order of degree_diff, 
    and in descending order of degree(u1).

    # Greedy Selection of Node Pairs:
    for each pair (degree_diff, u1, u2) in tot_degree_diff:
        if u1 and u2 are not matched:
            map u1 to u2 and update node_map, node_map_reverse, 
            matched_nodes1, matched_nodes2, H_nodes.
            create subgraph sg1 and sg2 from G1 and G2 
            using matched_nodes1 and matched_nodes2.
            check if every edge in sg1 has a corresponding edge in sg2 and 
            if every edge in sg2 has a corresponding edge in sg1:
                If any edge does not match, undo the match, 
                restore node_map, node_map_reverse, 
                matched_nodes1, matched_nodes2, and H_nodes.
                continue next mapping trial.

            If the match is successful, use BFS to extend the match:
                Initialize a queue with (u1, u2)
                while queue is not empty:
                    (v1, v2) ← dequeue queue
                    for each pair (k1, k2) in neighbors(v1)×neighbors(v2):
                        if k1 and k2 are not matched and 
                        |degree(k1) - degree(k2)|<=2:
                            map k1 to k2 and update node_map, node_map_reverse, 
                            matched_nodes1, matched_nodes2, and H_nodes.
                            create subgraph sg1 and sg2 from G1 and G2 
                            using matched_nodes1 and matched_nodes2.
                            check if every edge in sg1 has a corresponding edge in 
                            sg2 and if every edge in sg2 has a corresponding edge 
                            in sg1:
                                If any edge does not match, undo the match, 
                                restore node_map, node_map_reverse, 
                                matched_nodes1, matched_nodes2, and H_nodes.
                                continue next mapping trial.
    return length of H_nodes
\end{verbatim}
\end{tcolorbox}

\subsubsection{Traveling Salesman Problem}
\begin{tcolorbox}[colback=gray!10, colframe=black, rounded corners, boxrule=1.5pt, fontupper=\normalsize, left=2mm, right=2mm, top=1mm, bottom=1mm]
\begin{verbatim}
Function find_travelling_salesman_path(G):
    ans ← infinity
    n ← |V|
    for start in V do:
        length ← 0
        for each u in V do:
            State[u] ← 0
        v ← start
        State[v] ← 1
        for i ← 1 to n-1 do:
            select u in V that State[u] = 0 and G[v][u] is minimum
            length ← length + G[v][u]
            State[u] ← 1
            v ← u
        ans ← min(ans, length + G[v][start])
    return ans
\end{verbatim}
\end{tcolorbox}

\section{Executable Code Generated by PIE}
\label{code_generated_by_pie}
Furthermore, we present the optimal executable Python code generated by llama3-8b after multiple rounds of trials and errors using the aforementioned prompt.

\subsection{Common Neighbors}
\begin{mypython}
def common_neighbors(G, u, v):
    common_neighbors = []
    for i in range(len(G)):
        if i!= u and i!= v and (i in G[u] and i in G[v]):
            common_neighbors.append(i)
    return common_neighbors
\end{mypython}

\subsection{Connected Components}
\begin{mypython}
def connected_component_undirected(G):
    V = set(range(len(G)))
    V_prime = set()
    K = 1
    result = []
    while V!= V_prime:
        s = next((i for i in V if i not in V_prime), None)
        V_prime.add(s)
        C_K = [s]
        Q = [s]
        while Q:
            u = Q.pop(0)
            for v in G[u]:
                if v not in V_prime:
                    V_prime.add(v)
                    Q.append(v)
                    C_K.append(v)
        result.append(C_K)
        K += 1
    return K - 1
\end{mypython}

\subsection{Shortest Path}
\begin{mypython}
from collections import deque
def shortest_path_unweighted(G, u, v):
    Q = deque([(u, [u], 0)])
    V = set()
    while Q:
        node, path, dist = Q.popleft()
        if node == v:
            return dist
        V.add(node)
        for neighbor in G[node]:
            if neighbor not in V:
                Q.append((neighbor, path + [neighbor], dist + 1))
\end{mypython}

\subsection{Graph Diameter}
\begin{mypython}
from collections import deque
def graph_diameter(G):
    diameter = 0
    for i in range(len(G)):
        visited = set()
        distance = {j: float('inf') for j in range(len(G))}
        distance[i] = 0
        queue = deque([i])
        while queue:
            current_node = queue.popleft()
            visited.add(current_node)
            for neighbor in G[current_node]:
                if neighbor not in visited:
                    distance[neighbor] = min(distance[neighbor], distance[current_node] + 1)
                    queue.append(neighbor)
        diameter = max(diameter, max(distance.values()))
    return diameter
\end{mypython}

\subsection{Maximum Independent Set}
\begin{mypython}
def maximum_independent_set(G):
    n = len(G)
    degrees = [len(neighbors) for neighbors in G]
    sorted_nodes = sorted(range(n), key=lambda i: degrees[i])
    State = [0] * n
    IS = 0
    for v in sorted_nodes:
        if State[v] == 0:
            IS += 1
            State[v] = 1
            for u in G[v]:
                State[u] = 1
    return IS
\end{mypython}

\subsection{Minimum Vertex Cover}
\begin{mypython}
def minimum_vertex_cover(G):
    def count_cover(vertex):
        return sum(1 for neighbor in G[vertex] if (vertex, neighbor) in edges or (neighbor, vertex) in edges)

    edges = set((u, v) for u in range(len(G)) for v in G[u] if u < v)
    vertex_cover = set()
    while edges:
        max_coverage = 0
        best_vertex = -1
        for vertex in range(len(G)):
            if vertex not in vertex_cover and count_cover(vertex) > max_coverage:
                max_coverage = count_cover(vertex)
                best_vertex = vertex
        vertex_cover.add(best_vertex)
        for edge in list(edges):
            if best_vertex in edge:
                edges.remove(edge)
    return len(vertex_cover)
\end{mypython}

\subsection{Maximum Clique Problem}
\begin{mypython}
def maximum_clique(G):
    def bron_kerbosch_recursive(r, p, x):
        if not p and not x:
            nonlocal max_size
            max_size = max(max_size, len(r))
            C.append(r[:])
        for v in list(p):
            new_r = r + [v]
            new_p = p & set(neighbor for neighbor in G[v] if neighbor not in x)
            new_x = x & set(neighbor for neighbor in G[v] if neighbor not in r)
            bron_kerbosch_recursive(new_r, new_p, new_x)
            p.remove(v)
            x.add(v)
    max_size = 0
    C = []
    bron_kerbosch_recursive([], set(range(len(G))), set())
    return max_size
\end{mypython}

\subsection{Maximum Common Subgraph}
\begin{mypython}
import networkx as nx
from collections import deque
def maximum_common_subgraph(G1, G2):
    H_nodes = set()
    matched_nodes1 = set()
    matched_nodes2 = set()
    node_map = {}
    node_map_reverse = {}
    degree_diff = []
    for u1 in G1.nodes():
        for u2 in G2.nodes():
            degree_diff.append((abs(G1.degree(u1) - G2.degree(u2)), u1, u2))
    degree_diff.sort(key=lambda x: (x[0], -G1.degree(x[1])))
    for degree_diff, u1, u2 in degree_diff:
        if u1 not in matched_nodes1 and u2 not in matched_nodes2:
            node_map[u1] = u2
            node_map_reverse[u2] = u1
            matched_nodes1.add(u1)
            matched_nodes2.add(u2)
            H_nodes.add((u1, u2))
            sg1 = G1.subgraph(matched_nodes1)
            sg2 = G2.subgraph(matched_nodes2)
            if not nx.is_isomorphic(sg1, sg2):
                node_map.pop(u1)
                node_map_reverse.pop(u2)
                matched_nodes1.remove(u1)
                matched_nodes2.remove(u2)
                H_nodes.remove((u1, u2))
                continue
            queue = deque([(u1, u2)])
            while queue:
                v1, v2 = queue.popleft()
                for k1, k2 in zip(G1.neighbors(v1), G2.neighbors(v2)):
                    if k1 not in matched_nodes1 and k2 not in matched_nodes2 and abs(G1.degree(k1) - G2.degree(k2)) <= 2:
                        node_map[k1] = k2
                        node_map_reverse[k2] = k1
                        matched_nodes1.add(k1)
                        matched_nodes2.add(k2)
                        H_nodes.add((k1, k2))
                        sg1 = G1.subgraph(matched_nodes1)
                        sg2 = G2.subgraph(matched_nodes2)
                        if not nx.is_isomorphic(sg1, sg2):
                            node_map.pop(k1)
                            node_map_reverse.pop(k2)
                            matched_nodes1.remove(k1)
                            matched_nodes2.remove(k2)
                            H_nodes.remove((k1, k2))
                            continue
                        queue.append((k1, k2))
    return len(H_nodes)
\end{mypython}

\subsection{Traveling Salesman Problem}
\begin{mypython}
def travelling_salesman(G):
    n = len(G)
    ans = float('inf')
    for start in range(n):
        length = 0
        State = [0] * n
        v = start
        State[v] = 1
        for _ in range(n - 1):
            u = 0
            for i in range(n):
                if State[i] == 0 and (u == 0 or G[v][i] < G[v][u]):
                    u = i
            length += G[v][u]
            State[u] = 1
            v = u
        length += G[v][start]
        ans = min(ans, length)
    return ans
\end{mypython}

\section{Graph Prompt Example}
\subsection{Different Forms of Graph Structure Serialization}
\label{example_graph_structure}

\subsubsection{Edge List}
\begin{tcolorbox}[colback=gray!10, colframe=black, rounded corners, boxrule=1.5pt, fontupper=\normalsize, left=2mm, right=2mm, top=1mm, bottom=1mm]
graph edgelist is [(0, 1), (0, 2), (0, 3), (0, 4), (1, 2), (1, 3), (2, 3), (2, 4), (3, 4)]
\end{tcolorbox}

\subsubsection{Adj}
\begin{tcolorbox}[colback=gray!10, colframe=black, rounded corners, boxrule=1.5pt, fontupper=\normalsize, left=2mm, right=2mm, top=1mm, bottom=1mm]
0: [1, 2, 3, 4], 1: [0, 2, 3], 2: [0, 1, 3, 4], 3: [0, 1, 2, 4], 4: [0, 2, 3]
\end{tcolorbox}

\subsubsection{Adj NL}
\begin{tcolorbox}[colback=gray!10, colframe=black, rounded corners, boxrule=1.5pt, fontupper=\normalsize, left=2mm, right=2mm, top=1mm, bottom=1mm]
Node 0 is connected to nodes 1,2,3,4.
Node 1 is connected to nodes 0,2,3.
Node 2 is connected to nodes 0,1,3,4.
Node 3 is connected to nodes 0,1,2,4.
Node 4 is connected to nodes 0,2,3.
\end{tcolorbox}

\subsection{Different Forms of Pseudocode Prompt}
\label{different_pseudocode_prompt}
We use TSP as an example. The blue part indicates the differences.

\subsubsection{w/o pseudo}

\begin{tcolorbox}[colback=gray!10, colframe=black, rounded corners, boxrule=1.5pt, fontupper=\normalsize, left=2mm, right=2mm, top=1mm, bottom=1mm]

Given an undirected complete graph G where every pair of nodes is connected by an edge, and each edge has a weight, calculate the Traveling Salesman Problem (TSP) on the graph, which is the shortest cycle that visits each node exactly once and returns to the starting node. The Python code will receive the graph G, where the graph G is passed in the format of an adjacency matrix: G[i][j] stores the weight of edge (i, j). All elements are integers, and the graph G is numbered starting from 0.

Please complete the following Python code so that the function returns a value representing the solution to the Traveling Salesman Problem on the undirected graph G, i.e., the minimum edge weight sum of the cycle. def travelling\_salesman(G):
\end{tcolorbox}

\subsubsection{pseudo-NL}
\begin{tcolorbox}[colback=gray!10, colframe=black, rounded corners, boxrule=1.5pt, fontupper=\normalsize, left=2mm, right=2mm, top=1mm, bottom=1mm]

Given an undirected complete graph G where every pair of nodes is connected by an edge, and each edge has a weight, calculate the Traveling Salesman Problem (TSP) on the graph, which is the shortest cycle that visits each node exactly once and returns to the starting node. The Python code will receive the graph G, where the graph G is passed in the format of an adjacency matrix: G[i][j] stores the weight of edge (i, j). All elements are integers, and the graph G is numbered starting from 0.

\blue{The core idea of solving the Traveling Salesman Problem (TSP) in the provided code is to use a greedy approach to find a near-optimal path through all vertices of a complete graph. Here's a summary of the approach:

1. Initialization: Start by setting the initial answer (ans) to infinity, which will eventually hold the length of the shortest path found.

2. Iterate Over All Starting Vertices: For each vertex in the graph, treat it as the starting point of the path.

3. Greedy Path Construction:
    
    \quad 3.1. Initialize the path length to zero.
    
    \quad 3.2. Mark the starting vertex as visited and set it as the current vertex.
    
    \quad 3.3. For each of the remaining vertices (n-1 times), select the unvisited vertex that is closest to the current vertex (i.e., has the minimum edge weight to the current vertex).
    
    \quad 3.4. Add the edge weight to the path length and mark the selected vertex as visited.
    
    \quad 3.5. Move to the newly selected vertex and repeat the process until all vertices have been visited.

4. Complete the Cycle: After visiting all vertices, return to the starting vertex to complete the cycle. Add the edge weight from the last visited vertex back to the starting vertex to the path length.

5. Update the Best Path: Compare the total path length of the current cycle with the best path length found so far (ans). Update ans if the current path length is shorter.

6. Return the Result: After evaluating all possible starting vertices, return the shortest path length found (ans).
This approach is a heuristic method that aims to find a good (but not necessarily optimal) solution to the TSP by always choosing the nearest unvisited vertex at each step. The algorithm is efficient but does not guarantee the optimal solution, especially for larger graphs.}

Please complete the following Python code so that the function returns a value representing the solution to the Traveling Salesman Problem on the undirected graph G, i.e., the minimum edge weight sum of the cycle. def travelling\_salesman(G):
\end{tcolorbox}

\subsubsection{pseudo-Core}

\begin{tcolorbox}[colback=gray!10, colframe=black, rounded corners, boxrule=1.5pt, fontupper=\normalsize, left=2mm, right=2mm, top=1mm, bottom=1mm]

Given an undirected complete graph G where every pair of nodes is connected by an edge, and each edge has a weight, calculate the Traveling Salesman Problem (TSP) on the graph, which is the shortest cycle that visits each node exactly once and returns to the starting node. The Python code will receive the graph G, where the graph G is passed in the format of an adjacency matrix: G[i][j] stores the weight of edge (i, j). All elements are integers, and the graph G is numbered starting from 0.

\blue{The core idea of solving the Traveling Salesman Problem (TSP) in the provided code is to use a greedy approach to find a near-optimal path through all vertices of a complete graph. 

This approach is a heuristic method that aims to find a good (but not necessarily optimal) solution to the TSP by always choosing the nearest unvisited vertex at each step. The algorithm is efficient but does not guarantee the optimal solution, especially for larger graphs.}

Please complete the following Python code so that the function returns a value representing the solution to the Traveling Salesman Problem on the undirected graph G, i.e., the minimum edge weight sum of the cycle. def travelling\_salesman(G):
\end{tcolorbox}

\subsubsection{w/ pseudo}

\begin{tcolorbox}[colback=gray!10, colframe=black, rounded corners, boxrule=1.5pt, fontupper=\normalsize, left=2mm, right=2mm, top=1mm, bottom=1mm]

Given an undirected complete graph G where every pair of nodes is connected by an edge, and each edge has a weight, calculate the Traveling Salesman Problem (TSP) on the graph, which is the shortest cycle that visits each node exactly once and returns to the starting node. The Python code will receive the graph G, where the graph G is passed in the format of an adjacency matrix: G[i][j] stores the weight of edge (i, j). All elements are integers, and the graph G is numbered starting from 0.

\blue{The algorithm process can be described as follows:}
\begin{verbatim}
Function find_travelling_salesman_path(G):
    ans ← infinity
    n ← |V|
    for start in V do:
        length ← 0
        for each u in V do:
            State[u] ← 0
        v ← start
        State[v] ← 1
        for i ← 1 to n-1 do:
            select u in V that State[u] = 0 and G[v][u] is minimum
            length ← length + G[v][u]
            State[u] ← 1
            v ← u
        ans ← min(ans, length + G[v][start])
    return ans
\end{verbatim}
Please complete the following Python code so that the function returns a value representing the solution to the Traveling Salesman Problem on the undirected graph G, i.e., the minimum edge weight sum of the cycle. def travelling\_salesman(G):
\end{tcolorbox}

\end{document}